\documentclass{article} 
\usepackage{iclr2027_conference,times}

\usepackage{amsmath,amsfonts,bm}

\def\eqref#1{equation~\ref{#1}}

\def\1{\bm{1}}

\def\vc{{\bm{c}}}

\def\vh{{\bm{h}}}

\def\vw{{\bm{w}}}

\def\mA{{\bm{A}}}

\DeclareMathAlphabet{\mathsfit}{\encodingdefault}{\sfdefault}{m}{sl}
\SetMathAlphabet{\mathsfit}{bold}{\encodingdefault}{\sfdefault}{bx}{n}

\def\sY{{\mathbb{Y}}}

\newcommand{\E}{\mathbb{E}}

\newcommand{\R}{\mathbb{R}}

\usepackage{hyperref}
\usepackage{url}

\usepackage{booktabs}
\usepackage{graphicx}
\usepackage{subcaption}
\usepackage{amssymb}
\usepackage{bbm}
\usepackage{xcolor}
\usepackage{pifont}
\usepackage{makecell}
\usepackage{wrapfig}
\usepackage{microtype}

\usepackage{natbib}
\definecolor{citeblue}{RGB}{0,76,153}
\hypersetup{
    colorlinks=true,
    linkcolor=red,
    citecolor=citeblue,
    urlcolor=blue
}

\newcommand{\cmark}{\textcolor{green!60!black}{\ding{51}}}
\newcommand{\xmark}{\textcolor{red!70!black}{\ding{55}}}

\title{The Geometry of Knowledge Accessibility \\ in Large Language Models}

\author{Lihu Chen \\
Imperial College London\\
\texttt{lihu.chen@imperial.ac.uk} 
}

\iclrfinalcopy 
\begin{document}

\maketitle

\begin{abstract}
Large language models (LLMs) contain broad knowledge, but they cannot access all of it reliably. We study this problem through \emph{knowledge accessibility}, which describes whether the knowledge needed for a query can be recalled from the model.
We find that knowledge accessibility has a simple geometric structure in the model's representation of the query alone, before any generation.
More accessible queries are closer to a center in the representation space, while less accessible queries are farther away. This geometry reveals a \emph{knowledge boundary} that separates more accessible queries from less accessible ones. Accessibility consistently decreases with distance from the center, and this distance-based ordering transfers across datasets even when the centers differ. Controlled experiments further show that the centered geometry is more closely related to knowledge accessibility than to reasoning difficulty. The geometry also reveals when different interventions are useful. Query rewriting helps more for accessible queries, chain-of-thought reasoning helps more near the boundary, and retrieval gives larger gains beyond the boundary. These findings not only provide a new geometric view of how knowledge is organized in language models, but also suggest a useful pre-generation signal for adaptive inference.

\end{abstract}

\section{Introduction}

Large language models (LLMs) store a large amount of knowledge during
pre-training~\citep{petroni-etal-2019-language}, yet this knowledge is not equally accessible for every query.
This limitation is often described by the \emph{knowledge boundary}, which
separates queries that a model can answer from those that it
cannot~\citep{kadavath2022language,
li-etal-2025-knowledge-boundary}.
Recent work shows that this boundary can be estimated from the internal
states of a query before any token is generated
~\citep{ni-etal-2025-towards,
chen2026querylevel}.
Such generation-free signals are potentially useful for deciding how much additional computation or external information a query may require.

However, existing work often defines the knowledge boundary in terms of observed answer correctness.
This conflates knowledge accessibility with reasoning, as a model may access the required knowledge but still answer incorrectly due to reasoning failures~\citep{press-etal-2023-measuring, yang2026decoupling}.
Moreover, prior methods mainly ask whether knowledge is encoded in hidden states~\citep{dai-etal-2022-knowledge,chen2025identifying}, but they do not reveal how knowledge accessibility itself is organized in representation space.

In this work, we study the problem through \emph{knowledge accessibility}, which describes whether the knowledge required by a query is accessible from the model's parameters.
To isolate knowledge recall from reasoning, we focus on single-fact queries, such as ``\textit{The capital of France is?}'', whose answers  require one factual association rather than multi-step reasoning~\citep{petroni-etal-2019-language,press-etal-2023-measuring}.
This provides a controlled setting for studying the representation of accessibility.
We then learn the geometry of knowledge accessibility in query representations and ask a different question:
\begin{center}
\emph{What is the geometry of knowledge accessibility in an LLM's
representation space?}
\end{center}

We propose the
\emph{accessibility center hypothesis}.
Inspired by center-based structures in representation learning~\citep{wen2016discriminative,ruff2018deep}, we hypothesize that reliably accessible knowledge tends to occupy a localized region of representation space.
We model this region with an accessibility center and measure each query by
its distance from the center, with smaller distances indicating higher
knowledge accessibility.
This distance naturally defines a \emph{knowledge boundary} separating
regions of higher and lower accessibility.

Across two LLMs and four QA datasets, we find consistent support for this centered organization.
Simple centered geometries predict knowledge accessibility better than
competitive probing baselines on knowledge-focused tasks, and accessibility
consistently decreases with distance from the learned center.
In contrast, on reasoning-heavy tasks, a flexible MLP gains an advantage by
capturing additional reasoning signals, while the centered geometry
remains more closely associated with factual accessibility.
Controlled experiments further support this distinction between
accessibility and reasoning.

The geometric view also reveals several properties that are difficult to see
from correctness prediction alone.
Queries tend to be less stable in their correctness as they approach the
boundary, while their distance-based accessibility ordering transfers across
datasets even when the precise centers differ.
More importantly, the geometry suggests how different interventions benefit
the model. Query rewriting is most effective for accessible queries,
chain-of-thought reasoning becomes more useful near the boundary, and
retrieval provides larger gains beyond the boundary.
Thus, the knowledge boundary is not only a signal of whether a model
may need help, its geometry also provides information about what kind
of help is likely to be useful.
These results provide a geometric perspective on how knowledge is
organized and accessed in LLMs, and suggest that knowledge accessibility may
serve as a useful signal for adaptive and more efficient inference~\citep{NEURIPS2025_04185b5a,chen2026querylevel}.

\section{Knowledge Accessibility}

We conceptually break down query correctness as
\(C \approx f(K,R,O)\), where \(K\), \(R\), and \(O\) denote
\emph{knowledge accessibility}, \emph{reasoning capability}, and
\emph{other task-specific capabilities}.
That is, answering a query correctly requires both access to the needed knowledge
and the ability to use it appropriately.
This decomposition is motivated by prior work showing that access to relevant
knowledge and the ability to reason with that knowledge are distinct factors~\citep{press-etal-2023-measuring, yang2026decoupling}.
We treat knowledge accessibility ($K$) and reasoning capability ($R$) as the
primary factors of interest for query correctness, while $O$ captures other
task-specific factors whose influence is expected to be limited in our setting.

In this work, we focus on knowledge accessibility, which we define
as the ability to recall the factual knowledge required by a query from the
model's parameters.
To study knowledge accessibility separately from reasoning capability, we
focus on \emph{single-fact factual queries}. Answers of these queries mainly
depend on recalling a single target fact
~\citep{petroni-etal-2019-language,roberts-etal-2020-much}.
This setting allows us to isolate factual recall from multi-step
reasoning.

\subsection{Query Correctness and Factual Recall}
For a single-fact factual query $x$, let $\sY(x)$ denote the set of valid
answers and let $\hat{y}_{\pi}(x)$ be the model response under decoding
strategy $\pi$. We define
\begin{equation}
c_{\pi}(x) \triangleq \1[\hat{y}_{\pi}(x) \in \sY(x)] .
\label{eq:parametric-recall}
\end{equation}
In our single-fact setting, correctness mainly reflects whether the target fact
is successfully recalled. We therefore use $c_{\pi}(x)$ as a behavioral
observation of factual recall under decoding strategy $\pi$,
with correctness potentially varying across decoding strategies.

More generally, we view knowledge accessibility as the model's ability to
recall the required factual knowledge from its parameters under a given
elicitation and decoding setting. Low accessibility may reflect either knowledge that is not reliably stored or knowledge that is stored but not successfully accessed. This work does not distinguish between these two cases.

For more complex tasks, such as mathematical or multi-step reasoning,
correctness depends on more than knowledge accessibility. A model may have
access to the required knowledge but still fail because of reasoning or other
task-specific capabilities. 
This motivates our use of single-fact factual queries, where correctness provides a cleaner proxy for knowledge accessibility.

\subsection{A Geometric Hypothesis of Knowledge Accessibility}

Let $\vh_\ell(x) \in \R^{d}$ denote the hidden representation at the
final query token at layer $\ell$. We compute $\vh_\ell(x)$ from the query
alone, before generating any answer tokens. For notational simplicity, we
omit the layer index $\ell$ in the following.
We therefore ask whether knowledge accessibility is already reflected in
representation space before generation, and, if so, what geometric structure
it takes.

Prior work has mostly used linear probing to study this question.
These studies ask whether hidden representations can separate queries with
different model outcomes~\citep{xiao-etal-2025-analyzing, chen2026querylevel}. This shows that hidden states may contain information
about knowledge accessibility. However, it does not tell us how this information
is organized in representation space.

A simple center-based structure has been observed in learned representations.
In particular, representations with shared properties may become more
concentrated around a center in representation space
\citep{papyan2020prevalence,wu2024linguistic}.
We ask whether a similar effect also appears for knowledge accessibility.
Similar queries may develop a shared geometric structure during training.
Queries based on frequently seen knowledge may form a common center, while
queries based on rarely seen or unseen knowledge receive less shared training
signal and may lie farther from this center.

\paragraph{Accessibility Center Hypothesis.}
We propose that queries may be organized around an
\emph{accessibility center}, with distance from this center related to how
easily the model can recall the required knowledge. Specifically, we posit an
accessibility center $\vc \in \R^{d}$ and define the knowledge distance as
$
D(x) \triangleq d(\vh(x), \vc).
$
We hypothesize that factual recall becomes less likely as knowledge distance
increases. 
Let $X$ denote a query drawn from the data distribution.  Formally, 
\begin{equation}
    r_1 < r_2
    \quad\Longrightarrow\quad
    \E[c(X)\mid D(X)=r_1]
    \geq
    \E[c(X)\mid D(X)=r_2]
\end{equation}

The accessibility center should not be interpreted as a location where
individual facts are stored. Rather, it summarizes a region of representation
space associated with more reliable access to factual knowledge.

\section{Modeling the Geometry of Knowledge Accessibility}
\label{sec:geometry}

\subsection{Centered Accessibility Geometry}

Center-based structures have been widely used in representation learning~\citep{snell2017prototypical,ruff2018deep}.
For example, center loss encourages representations sharing the same label
to concentrate around a learned feature center \citep{wen2016discriminative}.
Motivated by this view, we instantiate the accessibility-center hypothesis
with a simple hyperspherical geometry. Specifically, we define
$
D(x) \triangleq \|\vh(x)-\vc\|_2,
$
so that points with the same knowledge distance lie on a hypersphere centered
at $\vc$.

This formulation makes an isotropic assumption. Distance from the center is
measured equally in all representation dimensions. Representations, however,
may vary along different dimensions at different scales~\citep{ethayarajh-2019-contextual}. We therefore relax
this assumption and consider an anisotropic extension,
\begin{equation}
D(x)
\triangleq
\bigl(\vh(x)-\vc\bigr)^\top
\mA
\bigl(\vh(x)-\vc\bigr),
\label{eq:ellipsoidal-distance}
\end{equation}
where $\mA \succeq 0$ is \emph{dimension weights}, which is diagonal and controls the relative scaling of
different representation dimensions. The resulting level
sets are hyperellipsoids rather than hyperspheres, while retaining the
assumption that accessibility is organized around a common center.

To test whether accessibility is organized around a center,
rather than merely encoded in the representation, we compare against two references. Linear probing is widely used to assess
whether a concept is encoded in representations. We therefore use the score
$\vw^\top\vh(x)+b$ as a linear baseline. We additionally consider an MLP,
with more parameters and nonlinear activations, which provides a more expressive reference. These models contrast a centered inductive bias with linear
separability and flexible nonlinear predictability, respectively, as
summarized in Table~\ref{tab:accessibility_hypotheses}.

\begin{table*}[t]
\centering

\begin{minipage}[t]{0.5\textwidth}
\centering
\tiny
\setlength{\tabcolsep}{4.5pt}
\renewcommand{\arraystretch}{0.95}

\begin{tabular}{@{}llll@{}}
\toprule
\textbf{Hypothesis}
& \textbf{Accessibility score}
& \textbf{Geometry}
& \textbf{Parameters} \\
\midrule
Linear
& $\vw^\top \vh+b$
& Hyperplane
& $\mathcal{O}(d)$ \\
Centered
& $-\lVert \vh-\vc\rVert_2^2$
& Hypersphere
& $\mathcal{O}(d)$ \\
Anisotropic centered
& $-(\vh-\vc)^\top\mA(\vh-\vc)$
& Hyperellipsoid
& $\mathcal{O}(d)$ \\
Flexible nonlinear
& $\operatorname{MLP}(\vh)$
& MLP reference
& $\mathcal{O}(dm)$ \\
\bottomrule
\end{tabular}

\captionof{table}{
Geometric hypotheses of knowledge accessibility in representation space.
Here, \(d\) is the query representation dimension and \(m\) is the MLP hidden dimension.}
\label{tab:accessibility_hypotheses}
\end{minipage}
\hfill
\begin{minipage}[t]{0.45\textwidth}
\centering
\tiny
\setlength{\tabcolsep}{4pt}
\renewcommand{\arraystretch}{1.03}

\begin{tabular}{@{}lccl@{}}
\toprule
\textbf{Dataset}
& \textbf{$K$}
& \textbf{$R$}
& \textbf{Task Type} \\
\midrule
TriviaQA
& High
& Low
& Factual QA \\
SciQ
& High
& Medium
& Science QA \\
MuSiQue
& High
& High
&  Multi-hop QA \\
GSM8K
& Medium
& High
& Math reasoning  \\
\bottomrule
\end{tabular}

\captionof{table}{
Decomposing answerability across benchmarks.
Correctness reflects knowledge accessibility ($K$)
and reasoning capability ($R$).
}
\label{tab:answerability_decomposition}
\end{minipage}

\end{table*}



\subsection{The Knowledge Boundary}
\label{sec:boundary}

Under the accessibility-center hypothesis, queries farther from the
accessibility center are less likely to support successful factual recall.
This radial ordering naturally induces a transition between regions of
higher and lower knowledge accessibility. We refer to this transition as
the \emph{knowledge boundary}.

Let $\tau$ denote a learned boundary radius and let $\epsilon>0$ control the
width of the transition region. Under this geometry, we define
three regions:
\begin{equation}
\begin{aligned}
D(x) &< \tau-\epsilon
&& \text{(\emph{interior})}, \
|D(x)-\tau| &\leq \epsilon
&& \text{(\emph{boundary})}, \
D(x) &> \tau+\epsilon
&& \text{(\emph{exterior})}.
\end{aligned}
\end{equation}

For subsequent analysis, we define the signed distance to the boundary as
$
\delta(x)
\triangleq
D(x)-\tau.
$
Accordingly, the three regimes can equivalently be expressed as
\begin{equation}
\begin{aligned}
\delta(x) &< -\epsilon
&& \text{(\emph{interior})}, \
|\delta(x)| &\leq \epsilon
&& \text{(\emph{boundary})}, \
\delta(x) &> \epsilon
&& \text{(\emph{exterior})}.
\end{aligned}
\end{equation}
Queries in the interior are expected to have higher knowledge accessibility, while those in the exterior have lower accessibility. The boundary marks an unstable transition region between the two.

\section{Evaluating the Accessibility Geometry}
\label{sec:experiments}

\subsection{Experimental Setup}

We evaluate two instruction-tuned language models,
\texttt{Llama-3.1-8B-Instruct}~\citep{grattafiori2024llama} and
\texttt{Qwen-2.5-14B-Instruct}~\citep{hui2024qwen2}, across four QA benchmarks with different
demands on knowledge accessibility and reasoning
(Table~\ref{tab:answerability_decomposition}).

TriviaQA~\citep{joshi-etal-2017-triviaqa} and SciQ~\citep{welbl-etal-2017-crowdsourcing} provide relatively clean settings for studying knowledge
accessibility, as correctness mainly reflects factual recall. MuSiQue~\citep{trivedi-etal-2022-musique} and
GSM8K~\citep{cobbe2021training} require more reasoning.
They let us test whether the centered geometry still works when correctness
depends less directly on knowledge access. In such settings, a more flexible nonlinear
model may become advantageous.
For each query, we obtain the observed answer using the greedy decoding policy \(\pi\) (Equation~\ref{eq:parametric-recall}), and use \texttt{Mistral-Large} to judge its correctness.

For each query, we extract the hidden state at the final query token from
the query alone, without generating any answer tokens. We use a
40\%/10\%/50\% train/validation/test split and use the same split for all
layers within each random seed. All results are averaged over five
random seeds.

For the layer-wise analysis, we fit each geometry independently at every
layer.  For all subsequent
analyses, we use the layer with the highest mean validation AUROC and keep this layer fixed. 
We set $\tau$ on the validation set as the distance
where the correctness probability is $0.5$.
We choose the boundary width $\epsilon$ such that the boundary region
corresponds to predicted correctness probabilities between $0.45$ and $0.55$.
Full training details and
hyperparameters are provided in Appendix~\ref{app:training_details}.

\begin{figure*}[t]
    \centering

    \begin{subfigure}[t]{0.235\textwidth}
        \centering
        \includegraphics[width=\linewidth,trim=8 8 8 8]
        {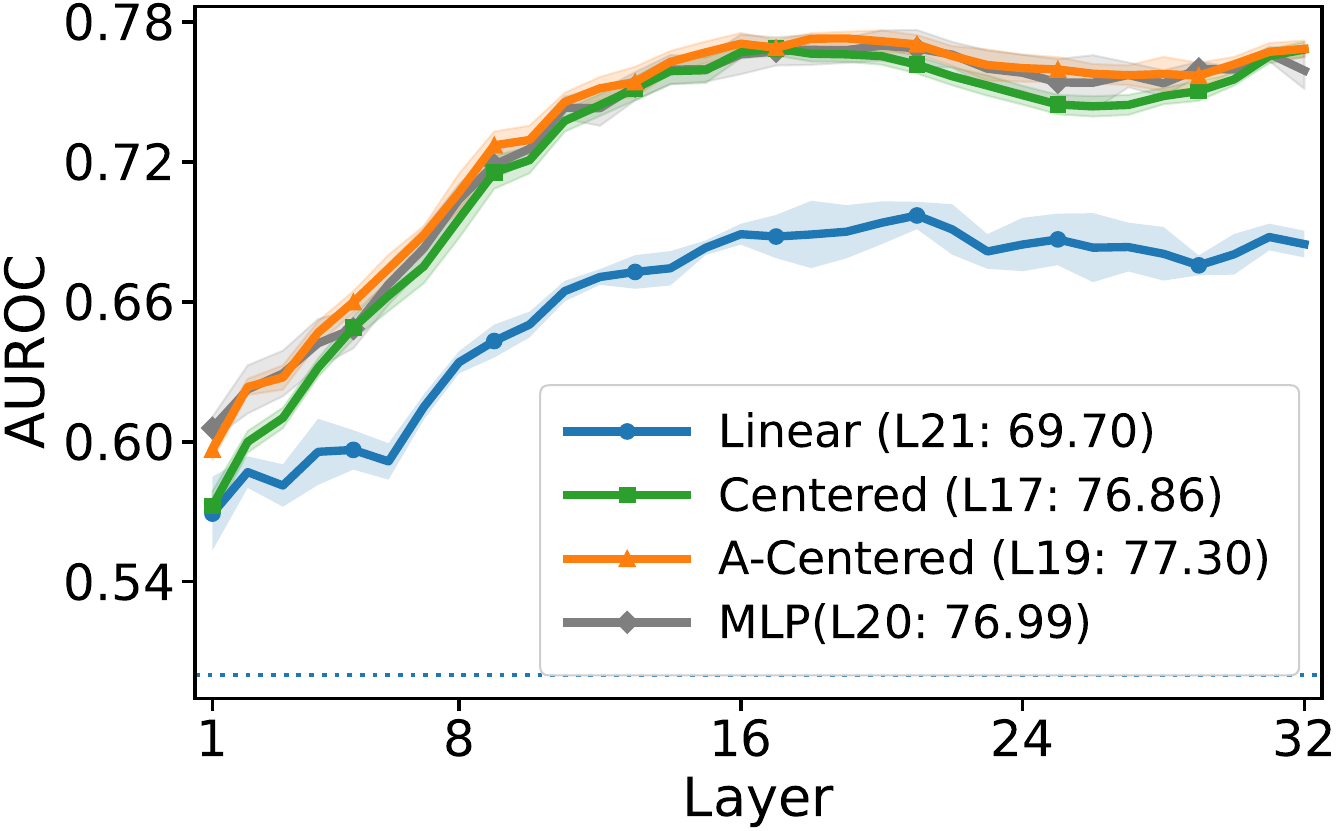}
        \caption{Llama-8B: TriviaQA}
        \label{fig:llama8b_triviaqa}
    \end{subfigure}
    \hfill
    \begin{subfigure}[t]{0.235\textwidth}
        \centering
        \includegraphics[width=\linewidth,trim=8 8 8 8]
        {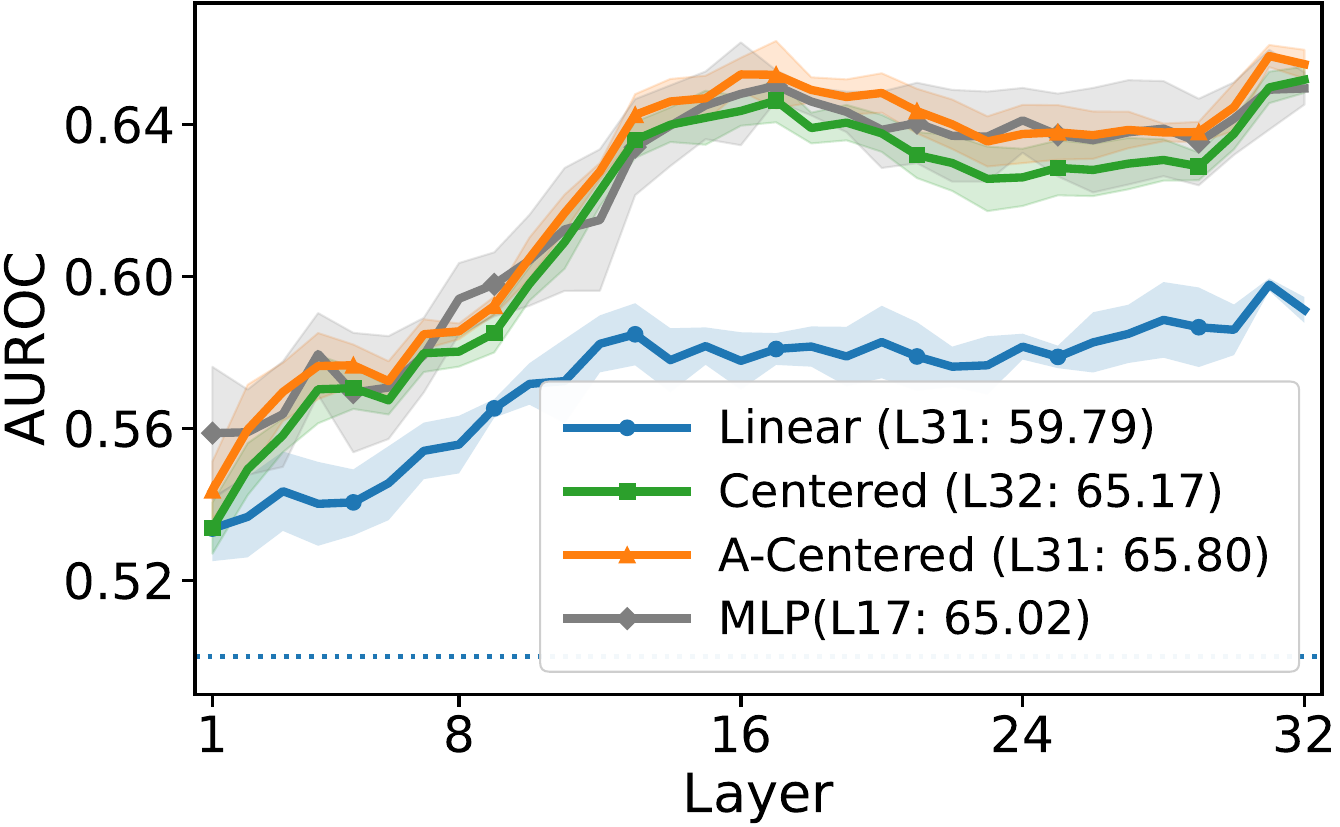}
        \caption{Llama-8B: SciQ}
        \label{fig:llama8b_sciq}
    \end{subfigure}
    \hfill
    \begin{subfigure}[t]{0.235\textwidth}
        \centering
        \includegraphics[width=\linewidth,trim=8 8 8 8]
        {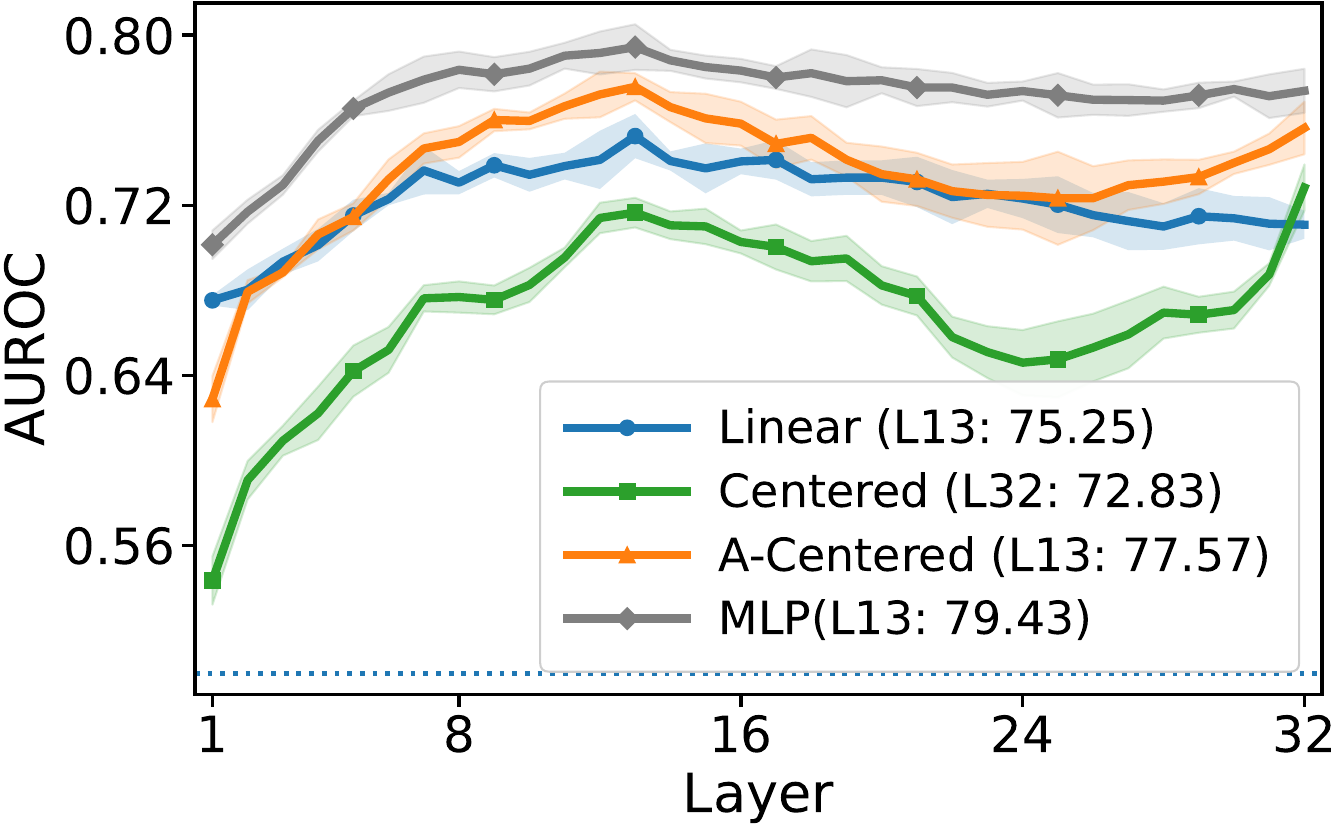}
        \caption{Llama-8B: MuSiQue}
        \label{fig:llama8b_musique}
    \end{subfigure}
    \hfill
    \begin{subfigure}[t]{0.235\textwidth}
        \centering
        \includegraphics[width=\linewidth,trim=8 8 8 8]
        {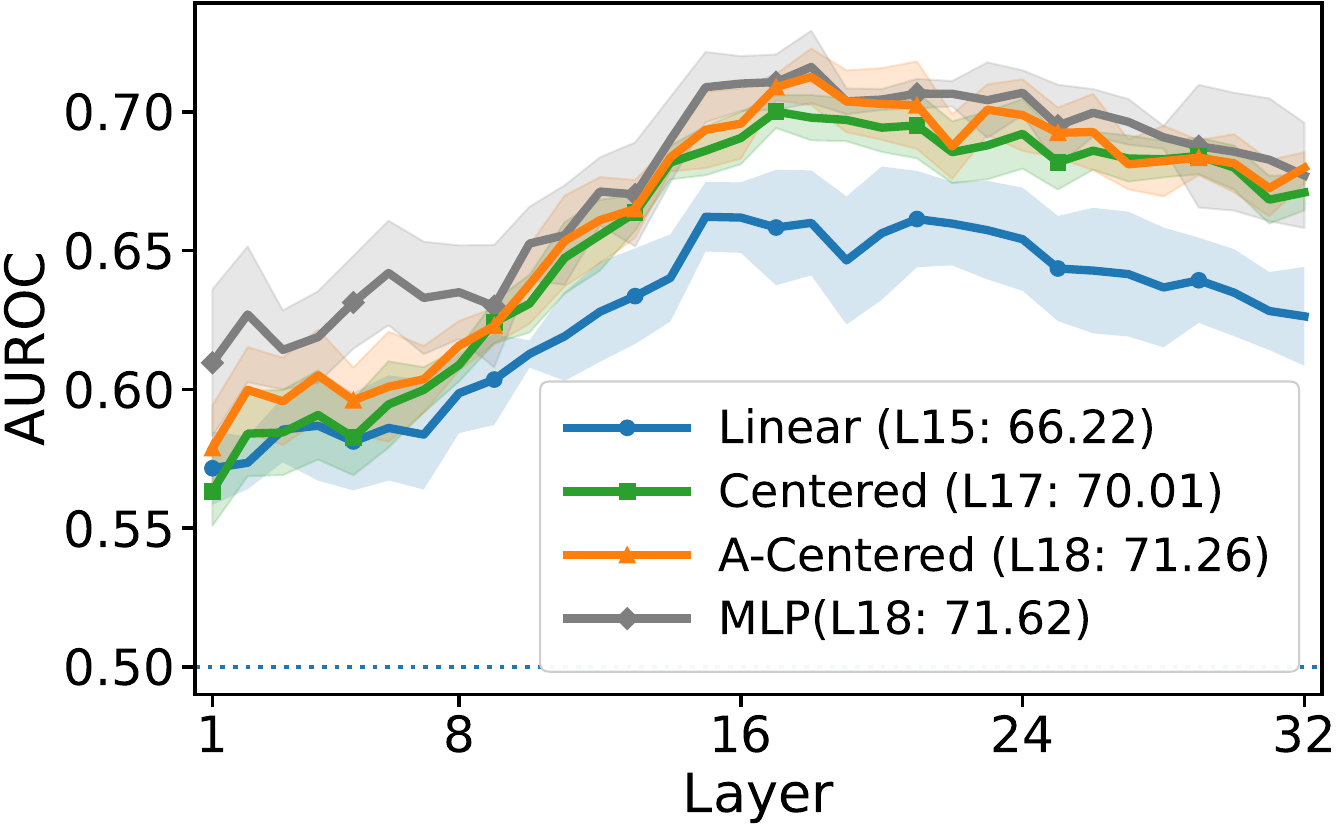}
        \caption{Llama-8B: GSM8K}
        \label{fig:llama8b_gsm8k}
    \end{subfigure}

    \vspace{0.2em}

    \begin{subfigure}[t]{0.235\textwidth}
        \centering
        \includegraphics[width=\linewidth,trim=8 8 8 8]
        {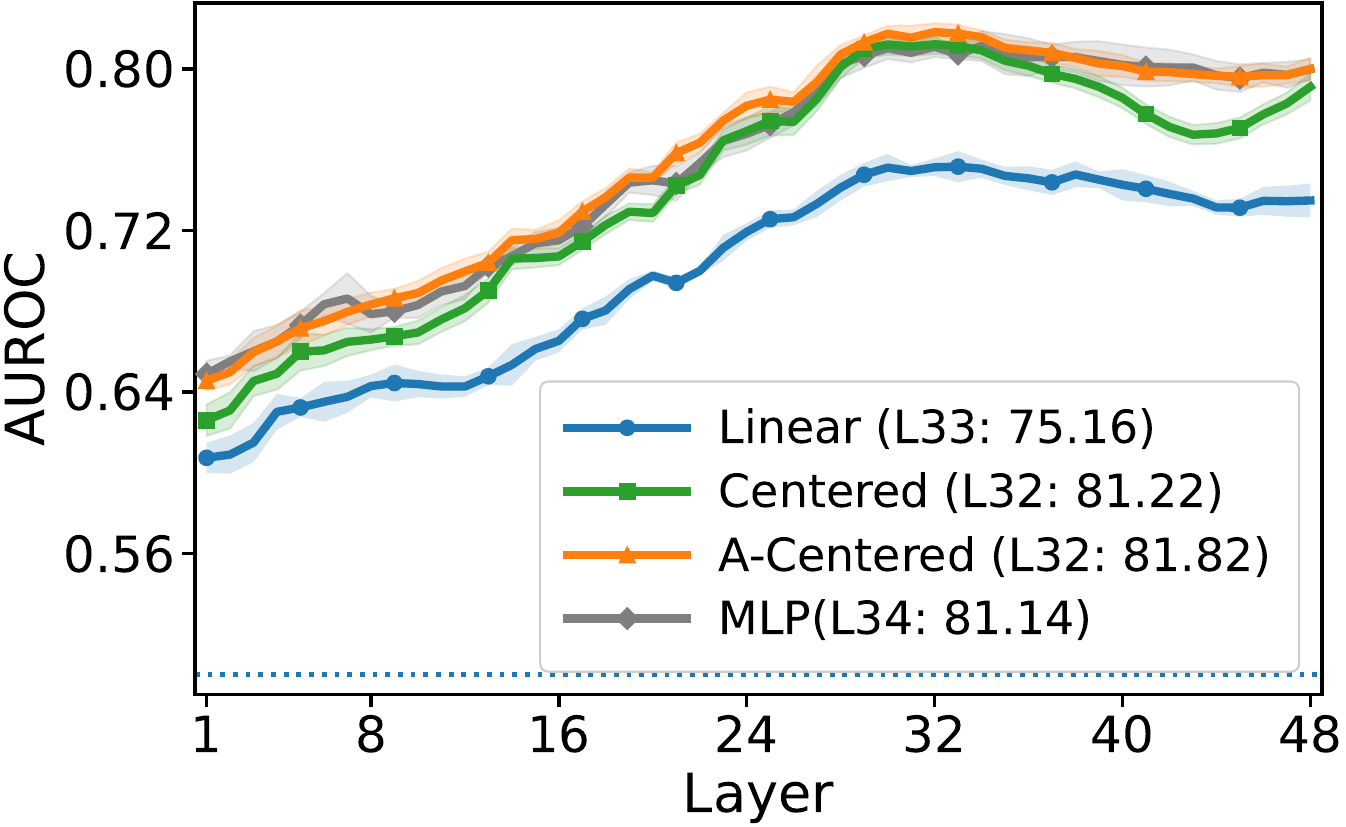}
        \caption{Qwen-14B: TriviaQA}
        \label{fig:qwen14b_triviaqa}
    \end{subfigure}
    \hfill
    \begin{subfigure}[t]{0.235\textwidth}
        \centering
        \includegraphics[width=\linewidth,trim=8 8 8 8]
        {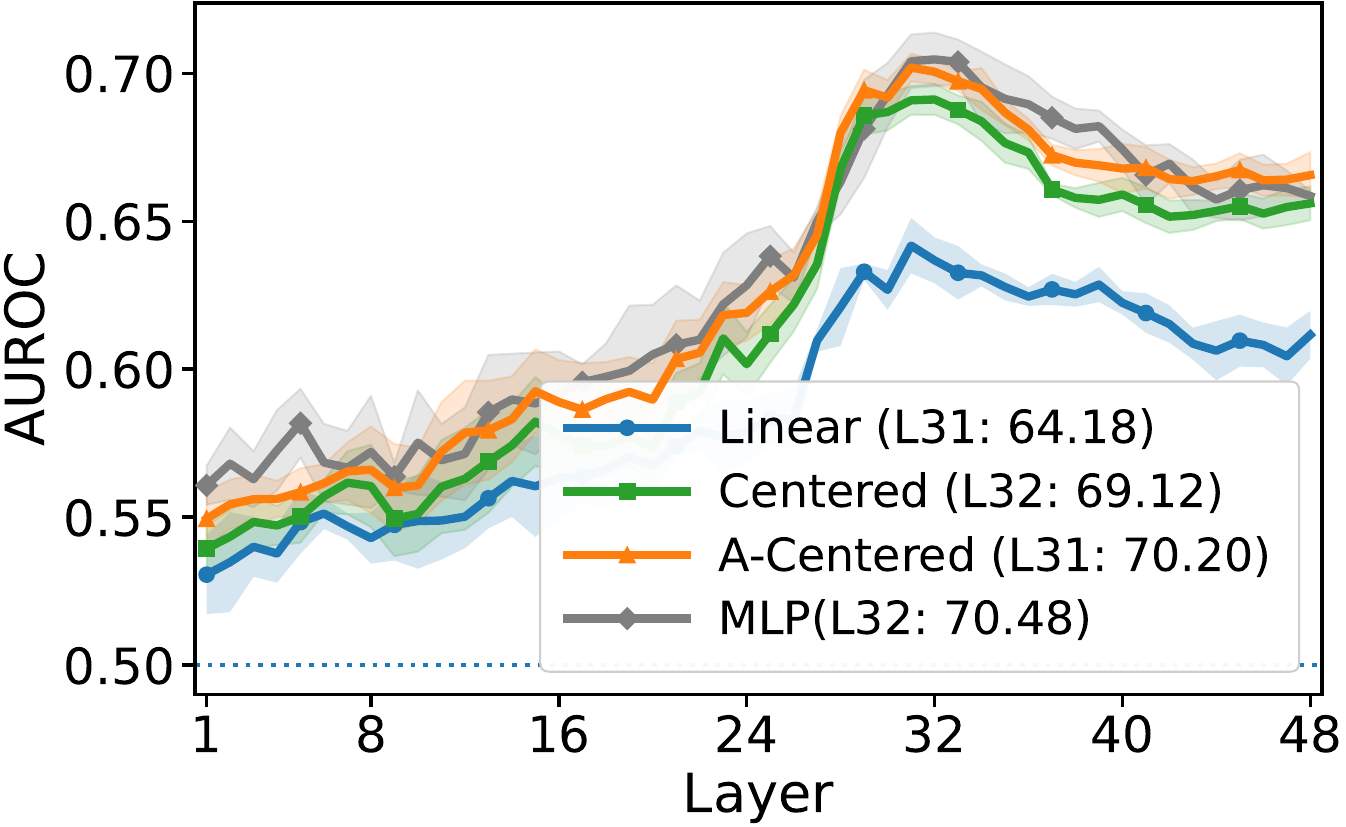}
        \caption{Qwen-14B: SciQ}
        \label{fig:qwen14b_sciq}
    \end{subfigure}
    \hfill
    \begin{subfigure}[t]{0.235\textwidth}
        \centering
        \includegraphics[width=\linewidth,trim=8 8 8 8]
        {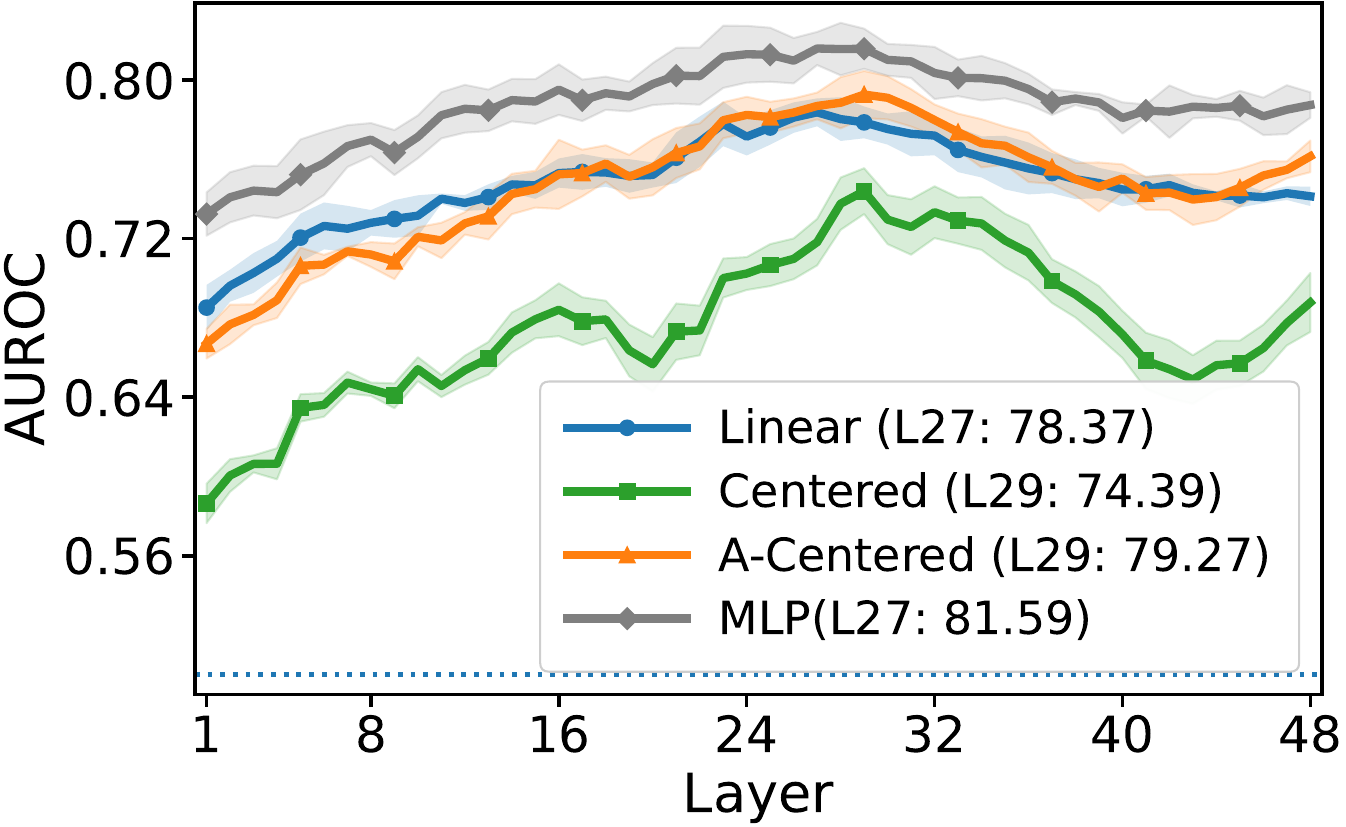}
        \caption{Qwen-14B: MuSiQue}
        \label{fig:qwen14b_musique}
    \end{subfigure}
    \hfill
    \begin{subfigure}[t]{0.235\textwidth}
        \centering
        \includegraphics[width=\linewidth,trim=8 8 8 8]
        {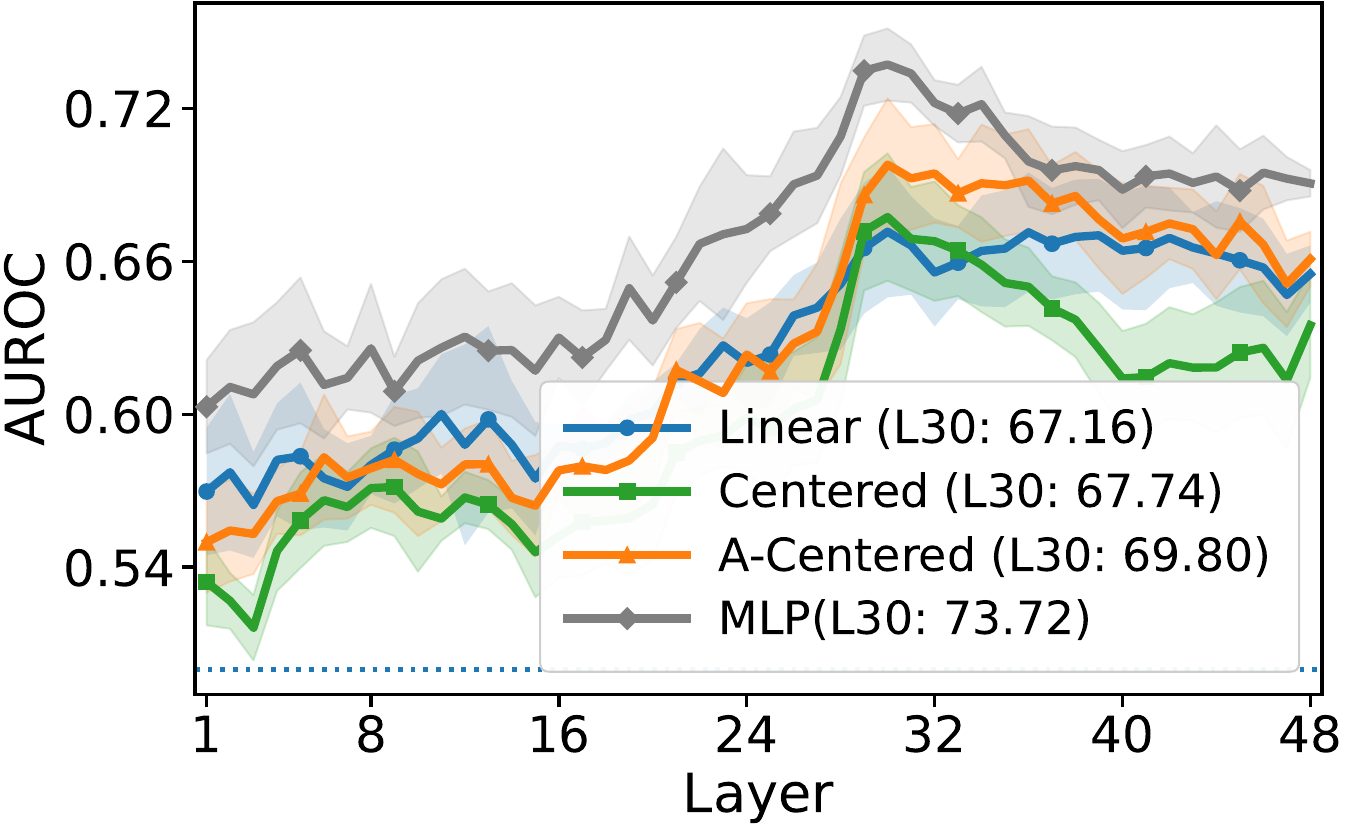}
        \caption{Qwen-14B: GSM8K}
        \label{fig:qwen14b_gsm8k}
    \end{subfigure}

    \caption{
    Layer-wise comparison of different geometry  hypotheses across four datasets and two language models.
    Peak performance is reported in the legend.
    }
    \label{fig:boundary_comparison_two_models}
\end{figure*}

\subsection{Comparing Knowledge Accessibility Geometries}
\label{sec:geometry-comparison}

Table~\ref{tab:accessibility_hypotheses} compares four hypotheses for how
answerability is organized in representation space: a linear geometry, an isotropic centered geometry, an anisotropic centered
geometry (\emph{A-Centered}), and a flexible nonlinear model (MLP).  Figure~\ref{fig:boundary_comparison_two_models}
evaluates these hypotheses layer by layer across the four benchmarks and two
language models.

On TriviaQA and SciQ, A-Centered is the strongest method on three of the
four model--dataset pairs. 
Notably, even the simple Centered geometry achieves performance close to the flexible MLP baseline.
In contrast, the Linear model performs substantially worse than the centered models.
Thus, when correctness is more directly driven by knowledge accessibility, a structured centered geometry is sufficient to match
or outperform a flexible nonlinear predictor. 
These results suggest that much of the accessibility signal can be captured by a simple centered geometry, rather than a complex nonlinear model.

On MuSiQue and GSM8K, where correctness depends more strongly on reasoning
capability, the flexible MLP performs best overall. This result is expected
because correctness on these tasks depends on both knowledge accessibility
and reasoning capability. Our hypothesis focuses on the geometric structure
of knowledge accessibility rather than correctness in general.
Even so, A-Centered remains competitive while using only $\mathcal{O}(d)$
parameters, compared with $\mathcal{O}(dm)$ for the MLP, as shown in Table~\ref{tab:accessibility_hypotheses}. This suggests that
the centered geometry still captures a large part of the correctness signal with a
much simpler model.

\begin{wrapfigure}{r}{0.35\textwidth}
    \centering
    \vspace{-0.6em}
    \includegraphics[width=\linewidth]
    {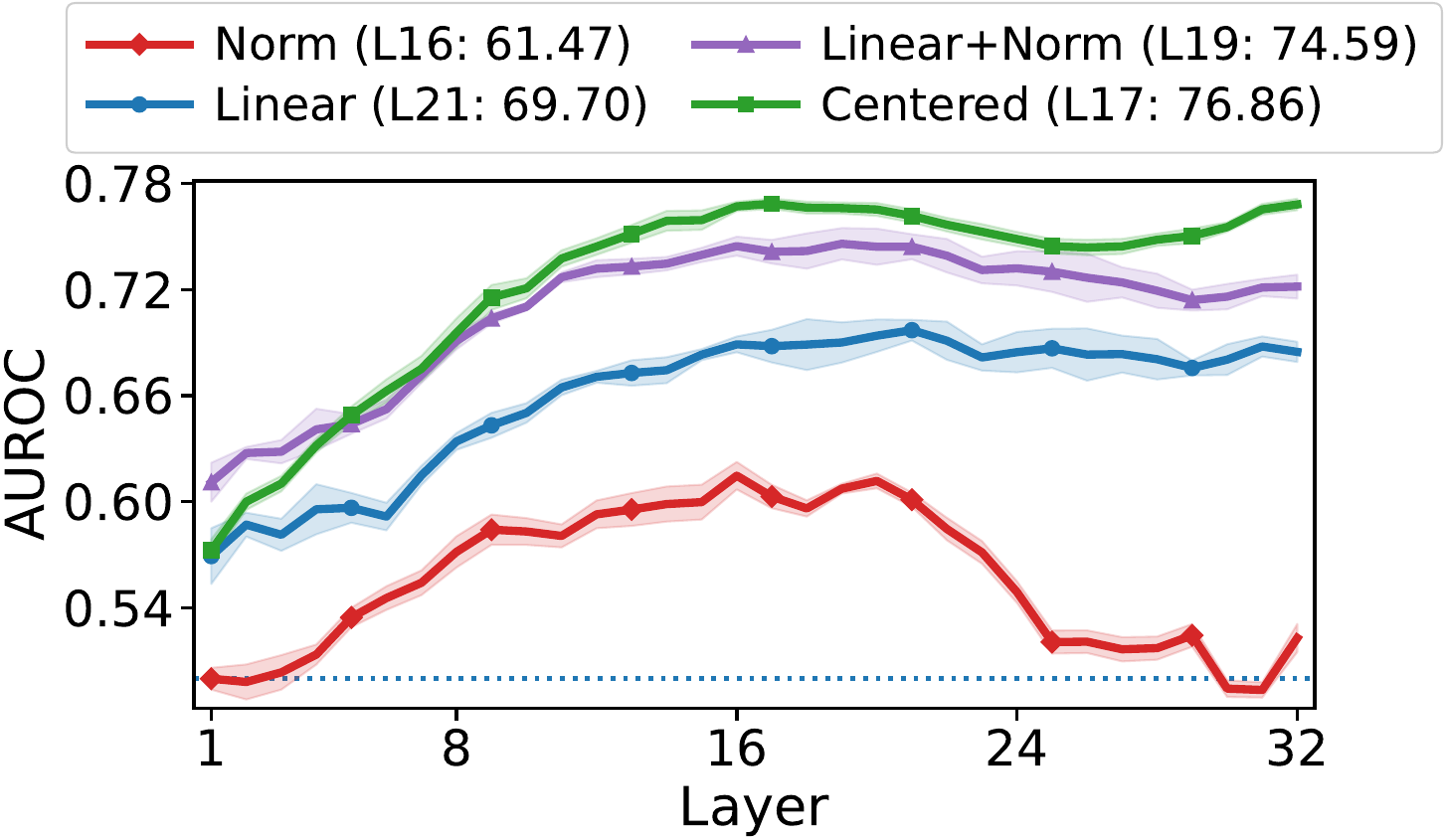}
    \caption{
        AUROC on TriviaQA with
        Llama-8B.
    }
    \label{fig:geometry-comparison}
    \vspace{-0.8em}
\end{wrapfigure}

One may ask whether Centered performs better than Linear simply because it
uses the squared norm of the representation, rather than because of the
centered structure itself. The Centered score can be expanded as
$
-\|\vh-\vc\|_2^2
=
2\vc^\top\vh
-
\|\vh\|_2^2
-
\|\vc\|_2^2,
$
which contains both a linear term and a squared norm term. We therefore
introduce \emph{Linear+Norm},
$
s(\vh)=\vw^\top\vh+\alpha\|\vh\|_2^2+b,
$
which learns these terms freely without enforcing a centered structure.
For AUROC, Centered depends on $d$ center parameters, while Linear+Norm
depends on $d$ linear weights and one norm coefficient $\alpha$, giving it
at least as much effective capacity.

As shown in Figure~\ref{fig:geometry-comparison}, Linear+Norm improves over
Linear but remains below Centered. This suggests that the benefit of
Centered is not explained by the additional norm term or greater model
capacity, but is consistent with a useful centered inductive bias.
We further analyze the learned coefficient $\alpha$ in
Appendix~\ref{app:centered_bias}, which shows that the 
Linear+Norm model also tends to learn a centered
structure.

\subsection{Separating Knowledge Accessibility from Reasoning}

\begin{wraptable}{r}{0.48\textwidth}
    \centering
    \small
    \setlength{\tabcolsep}{3.2pt}
    \renewcommand{\arraystretch}{0.98}
    \begin{tabular}{@{}lcccc@{}}
        \toprule
        \textbf{Group} & \textbf{K / R} & $n$ & \textbf{Dist.} & \textbf{Position} \\
        \midrule
        Success
        & \cmark{} / \cmark{} & 312 & 7.85 & Interior \\
        Reasoning Failure
        & \cmark{} / \xmark{} & 311 & 5.33 & Interior \\
        Knowledge Failure
        & \xmark{} / \xmark{} & 281 & 1.05 & Exterior \\
        Recovered Success
        & \xmark{} / \cmark{} & 96 & 5.29 & Interior \\
        \bottomrule
    \end{tabular}
    \caption{MQuAKE groups by absolute distance to the knowledge boundary, where K and R denote knowledge accessibility (\emph{single-hop}) and reasoning (\emph{multi-hop}) success, respectively.}
    \label{tab:mquake_boundary_position}
    \vspace{-0.7em}
\end{wraptable}

To further test more directly whether the centered geometry captures knowledge accessibility rather than reasoning capability,  we construct a controlled evaluation on MQuAKE~\citep{zhong-etal-2023-mquake}.
MQuAKE provides multi-hop questions together with their single-hop subquestions, which allows us to separate knowledge accessibility from reasoning capability. The single-hop questions test whether the individual facts needed for the question can be recalled, while the multi-hop question tests whether the model can use reasoning to combine these facts and produce the final answer. Thus, if the single-hop questions are answered correctly but the multi-hop question is answered incorrectly, the failure is more likely due to reasoning rather than missing knowledge. In contrast, failure on the required single-hop facts provides evidence of limited knowledge accessibility. We then apply the A-Centered geometry learned on TriviaQA directly to MQuAKE without relearning, which makes MQuAKE an independent test of the learned geometry.
(see Appendix~\ref{app:mquake_control} for details).

As shown in Table~\ref{tab:mquake_boundary_position}, geometric position is
more closely aligned with knowledge accessibility than with final correctness.
Both \emph{Success} and \emph{Reasoning Failure} cases have accessible
single-hop facts, and both remain inside the boundary, even though
the latter fail on the final multi-hop question.
In contrast, \emph{Knowledge Failure} cases, where at least one required fact
cannot be recalled, move into the exterior.
This pattern suggests that queries with similar knowledge accessibility remain close in distance despite different reasoning outcomes.
Interestingly, \emph{Recovered Success} cases also remain inside the boundary.
This suggests that failure on an isolated single-hop probe does not always mean
that the required knowledge is inaccessible.
Some of these accessibility failures may instead reflect sensitivity to prompting or decoding, which we study later.

\begin{figure*}[t]
    \centering

    \begin{subfigure}[t]{0.25\textwidth}
        \centering
        \includegraphics[width=\linewidth]
        {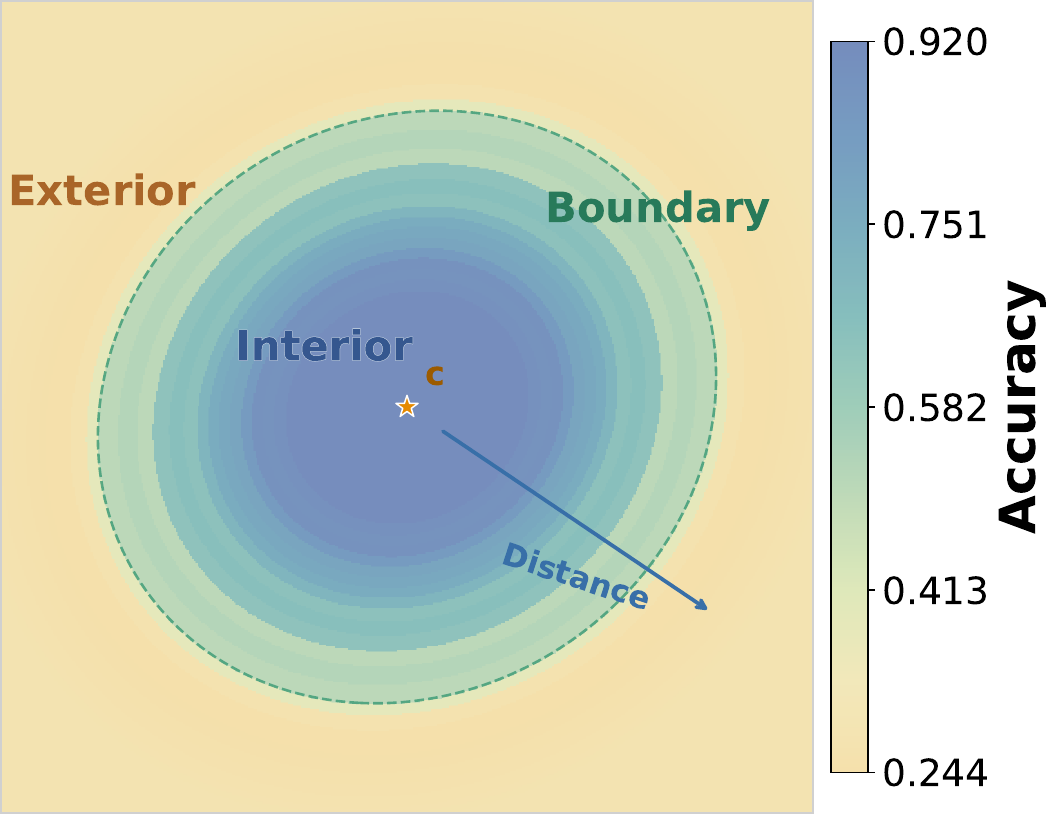}
        \caption{Accuracy vs. distance}
        \label{fig:knowledge_landscape}
    \end{subfigure}
    \hfill
    \begin{subfigure}[t]{0.3\textwidth}
        \centering
        \includegraphics[width=\linewidth]
        {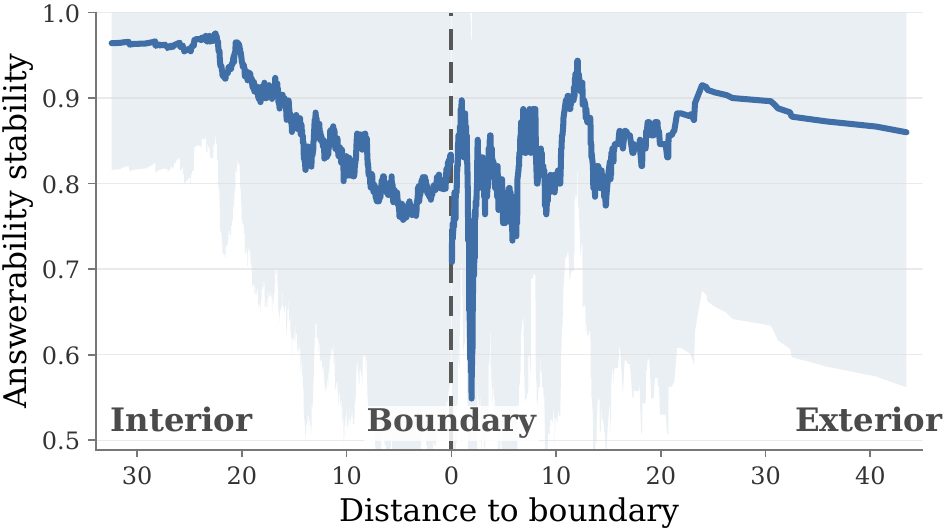}
        \caption{Correctness stability}
        \label{fig:answerability_stability}
    \end{subfigure}
    \hfill
    \begin{subfigure}[t]{0.3\textwidth}
        \centering
        \includegraphics[width=\linewidth]
        {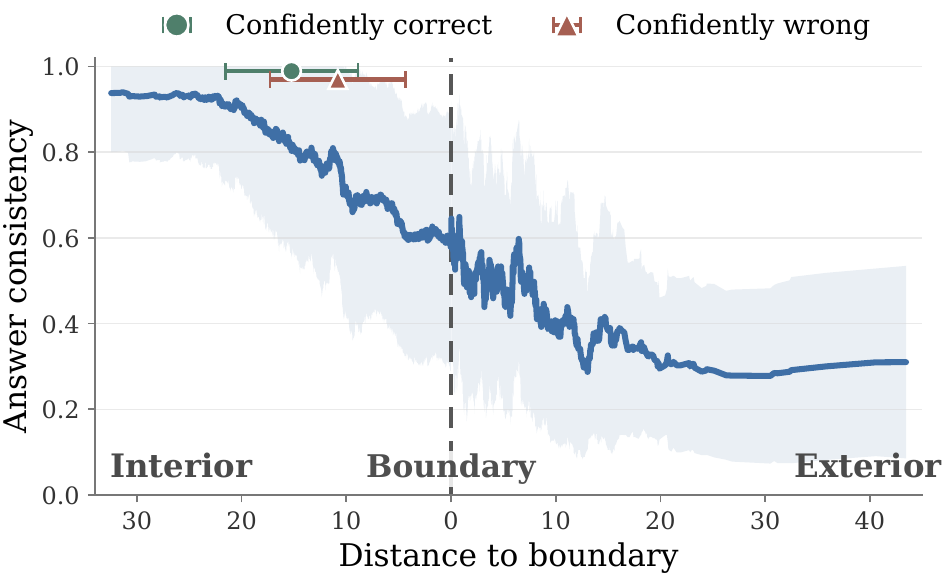}
        \caption{Answer consistency}
        \label{fig:answer_consistency}
    \end{subfigure}

    \vspace{-0.3em}
    \caption{
    Behavior of knowledge accessibility for Llama-8B on Trivia QA.
    (a) Accessibility decreases with distance from the center.
    (b) Answerability is least stable near the boundary.
    (c) Consistency decreases with distance from the center. Results for Qwen14B are provided in \ref{fig:qwen14b_knowledge_boundary_properties} in Appendix.
    }
    \label{fig:knowledge_boundary_properties}
    \vspace{-0.5em}
\end{figure*}

\subsection{Behavior of Knowledge Accessibility}
\label{sec:boundary_behavior}
We next test whether the learned geometry captures meaningful behavior beyond binary correctness.
In particular, we study how knowledge accessibility
changes with distance and how stable model behavior is around the learned
boundary.

\paragraph{Accessibility Monotonicity.}
We next visualize how knowledge accessibility changes across the learned
geometry. Figure~\ref{fig:knowledge_landscape} shows a low-dimensional
projection of the hidden representations and the learned boundary on
TriviaQA. Since TriviaQA mainly tests factual recall, accuracy provides a
 proxy for knowledge accessibility. Queries closer to the
center tend to have higher accessibility, while accessibility decreases
toward and beyond the boundary. This pattern is consistent with the learned
accessibility structure.

\paragraph{Correctness Stability.}
We next study how stable correctness is around the learned boundary.
For each query $x$, we sample $N$ responses under a fixed stochastic
decoding strategy $\pi$. Let $c_{\pi}^{(i)}(x)$ denote the correctness of
the $i$-th sampled response. We define
\begin{equation}
\hat{p}_{\pi}(x)
=
\frac{1}{N}\sum_{i=1}^{N} c_{\pi}^{(i)}(x),
\qquad
S_{\pi}(x)
\triangleq
\left|2\hat{p}_{\pi}(x)-1\right|.
\label{eq:correctness-stability}
\end{equation}
Here, $S_{\pi}(x)\in[0,1]$ is the correctness stability. A value close to $1$ means that repeated
generations are consistently correct or consistently incorrect, while a
value close to $0$ means that correctness varies across generations.

Figure~\ref{fig:answerability_stability} shows a U-shaped pattern of stability around the
learned boundary. Queries deep in the interior are stably answered correctly.
Stability decreases toward the boundary, where the same query can be answered
correctly in some generations and incorrectly in others. Beyond the boundary,
stability increases again, but now mainly reflects repeated failure. Together
with the decrease in accuracy, this gives the following behavioral pattern:
\text{\textcolor{blue}{stable access}}
$\;\longrightarrow\;$
\text{\textcolor{orange}{unstable access}}
$\;\longrightarrow\;$
\text{\textcolor{red}{stable failure}}.
The boundary therefore marks a region where access to the required knowledge
is least stable. Since repeated generations are not used to learn the
boundary, this provides independent evidence for the learned
geometry.

\paragraph{Answer Consistency.}
Figure~\ref{fig:answer_consistency} provides a complementary
view through answer consistency. 
On average, repeated answers become less consistent as distance increases, which suggests that knowledge access becomes less reliable farther from the accessibility center.
Some uncertainty methods rely on answer consistency across repeated generations. A limitation of such signals is that they cannot identify \emph{confidently wrong} samples~\citep{farquhar2024detecting}, while the centered geometry provides richer signals.

\begin{figure}[t]
\centering
\includegraphics[width=0.88\linewidth]{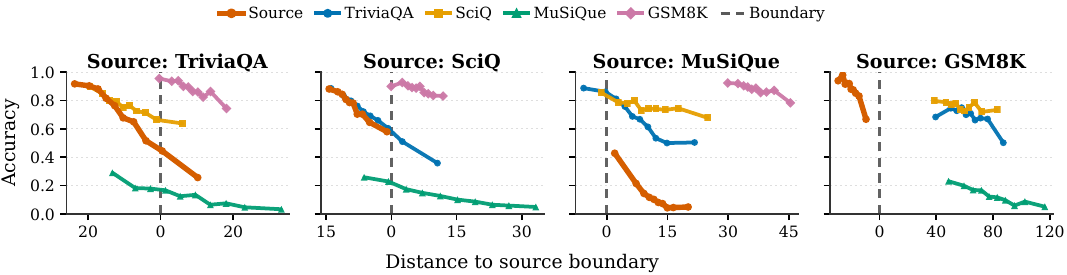}
\caption{
Transferability of the distance and accuracy relation for Llama-8B. Accuracy generally decreases with distance across target datasets. Results for Qwen-14B are provided in Figure~\ref{fig:qwen14b_cross-dataset-transfer} in Appendix.
}
\label{fig:cross-dataset-transfer}
\end{figure}

\subsection{Transferability of Accessibility Distance}
\label{sec:geometric-transferability}

We next study whether distance to an accessibility center learned on one
dataset can transfer to other datasets. We do not require the learned center to be the same across datasets. 
Instead, we ask
whether distance under a geometry learned on one dataset can still preserve
the accessibility ordering of queries from another dataset.
Full experimental
details are provided in Appendix~\ref{app:boundary-transfer}.

For each source dataset, we learn an accessibility center $\vc$ and its dimension
weights $\mA$, and apply them directly to the target datasets without
relearning. As shown in Figure~\ref{fig:cross-dataset-transfer}, task
performance generally decreases as the distance from the source accessibility
center increases across different source-target pairs. This transfer is
stronger on datasets where correctness mainly reflects knowledge accessibility.
On reasoning datasets (e.g., GSM8K), the distance still preserves a useful ordering,
but the relation is weaker. Overall, queries farther from the source center tend to
have lower knowledge accessibility.

However, the accessibility
centers $\vc$ and the dimension weights $\mA$ do not transfer across datasets, as shown in Figure~\ref{fig:geometry_heatmaps} in the Appendix. This suggests that
different types of task have their own accessibility centers. Even so, the monotonic relation between distance and knowledge
accessibility remains consistent across datasets. Queries farther from the
source center still tend to have lower knowledge accessibility.

\section{Knowledge Accessibility Interventions}
\label{sec:model-behavior}

So far, we have shown that distance is related to knowledge accessibility. Importantly, this distance is computed
solely from the query representation, before any answer tokens are
generated. For a query $x$, we extract its hidden representation $\vh(x)$ and
measure its distance to the learned accessibility center,
$D(x)=d(\vh(x),\vc)$.

We next ask whether effective knowledge accessibility can be improved at inference time, without retraining or modifying the model parameters. 
We consider three complementary interventions that operate through different mechanisms. (1) \emph{Query rewriting} changes the wording of the query without changing or adding new
information, (2) \emph{chain-of-thought (CoT)} provides additional computation, and
(3) \emph{retrieval-augmented generation (RAG)} provides external information.

\begin{figure*}[t]
    \centering

    \begin{subfigure}[t]{0.315\textwidth}
        \centering
        \includegraphics[width=\linewidth]
        {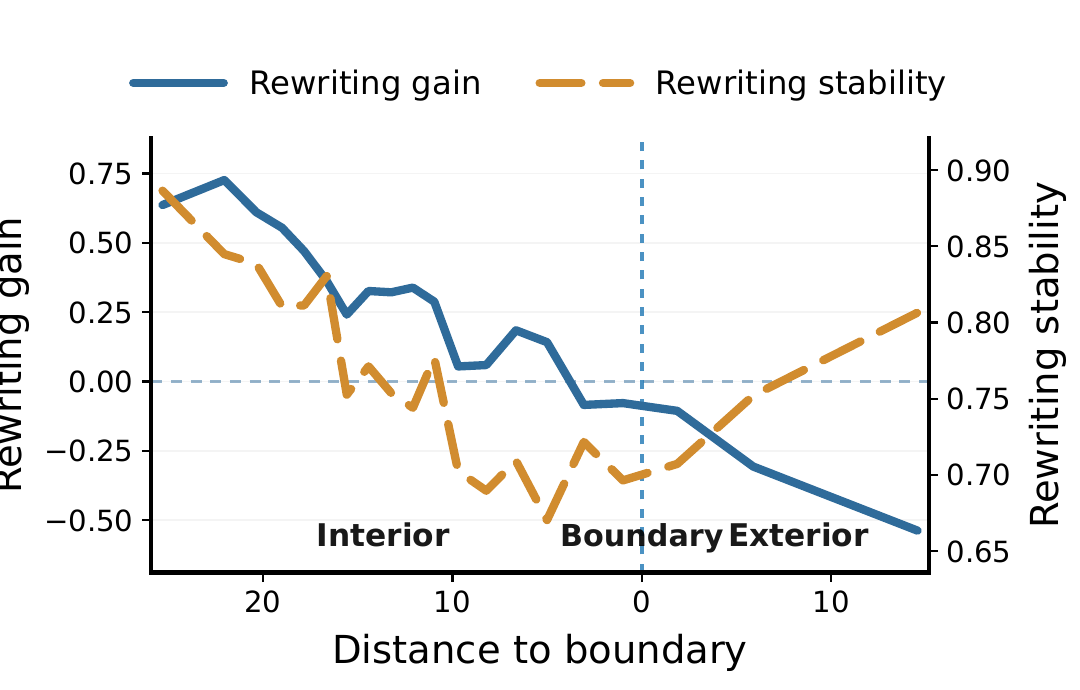}
        \caption{Llama-8B: Rewriting}
        \label{fig:llama_para_gain}
    \end{subfigure}
    \hfill
    \begin{subfigure}[t]{0.315\textwidth}
        \centering
        \includegraphics[width=\linewidth]
        {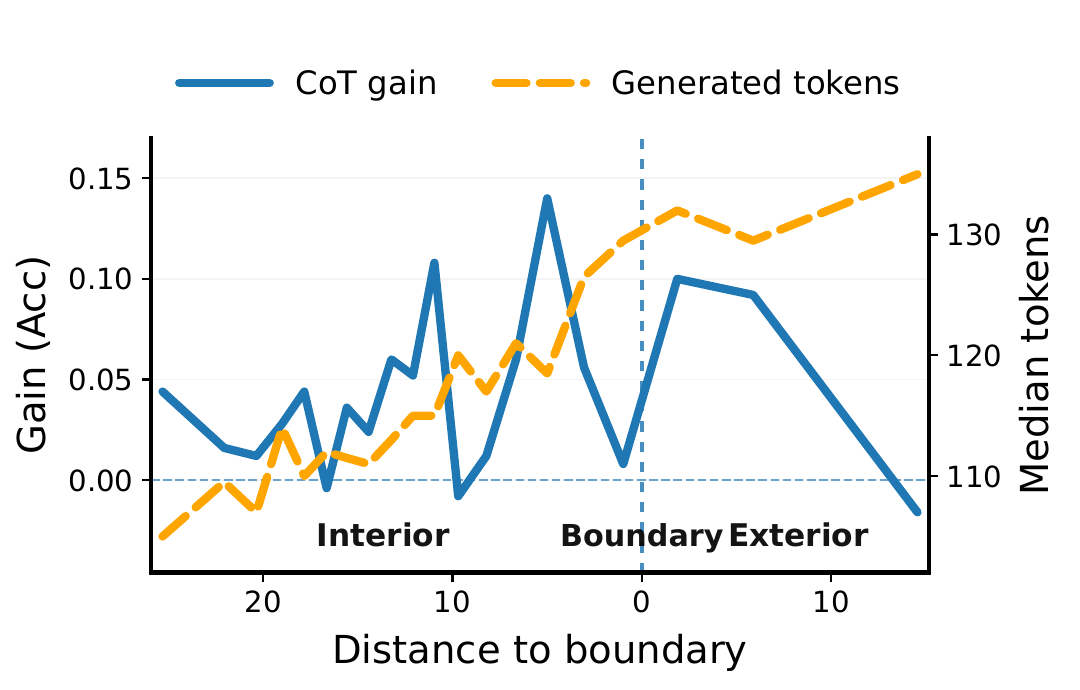}
        \caption{Llama-8B: CoT}
        \label{fig:llama_cot_gain}
    \end{subfigure}
    \hfill
    \begin{subfigure}[t]{0.315\textwidth}
        \centering
        \includegraphics[width=\linewidth]
        {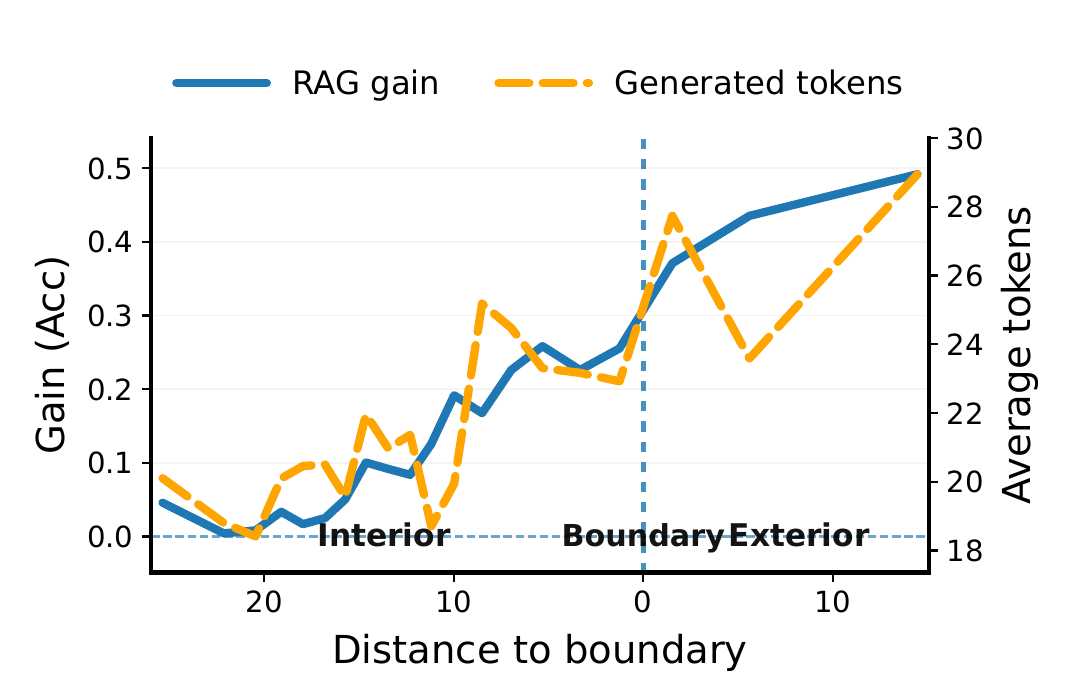}
        \caption{Llama-8B: RAG}
        \label{fig:llama_rag_gain}
    \end{subfigure}

    \vspace{0.5em}

    \begin{subfigure}[t]{0.315\textwidth}
        \centering
        \includegraphics[width=\linewidth]
        {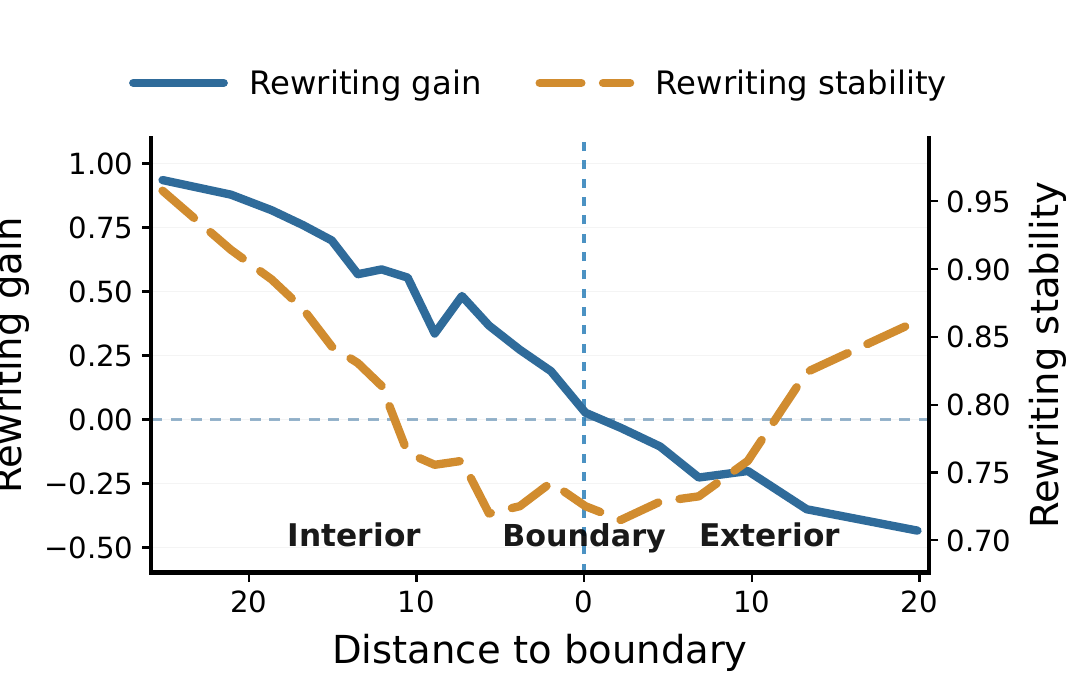}
        \caption{Qwen-14B: Rewriting}
        \label{fig:qwen_para_gain}
    \end{subfigure}
    \hfill
    \begin{subfigure}[t]{0.315\textwidth}
        \centering
        \includegraphics[width=\linewidth]
        {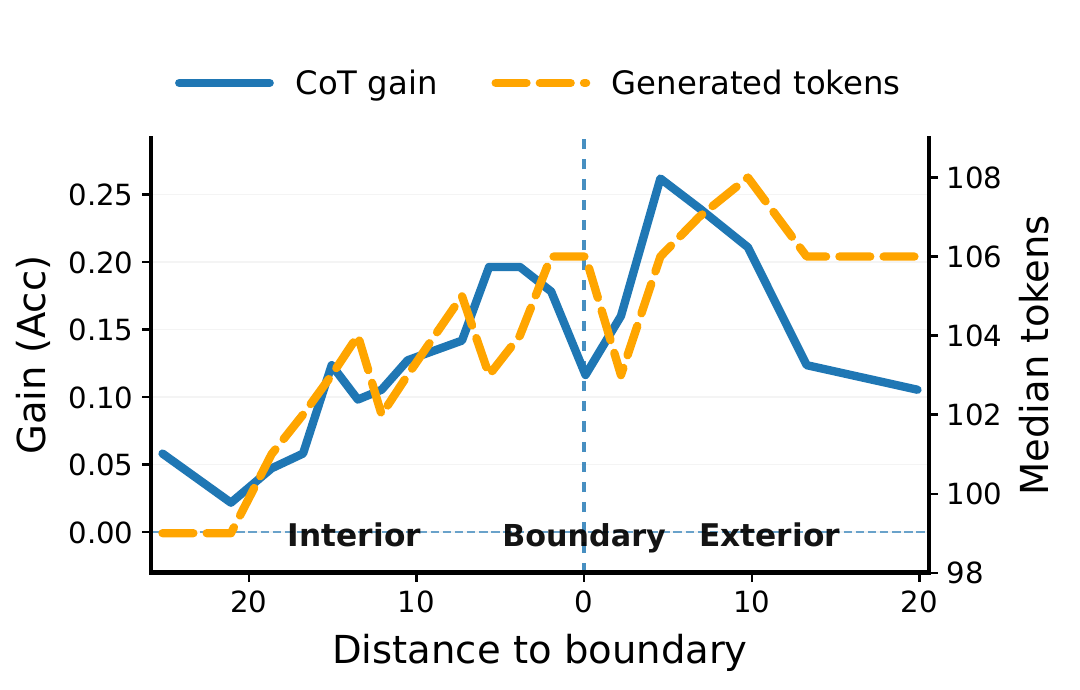}
        \caption{Qwen-14B: CoT}
        \label{fig:qwen_cot_gain}
    \end{subfigure}
    \hfill
    \begin{subfigure}[t]{0.315\textwidth}
        \centering
        \includegraphics[width=\linewidth]
        {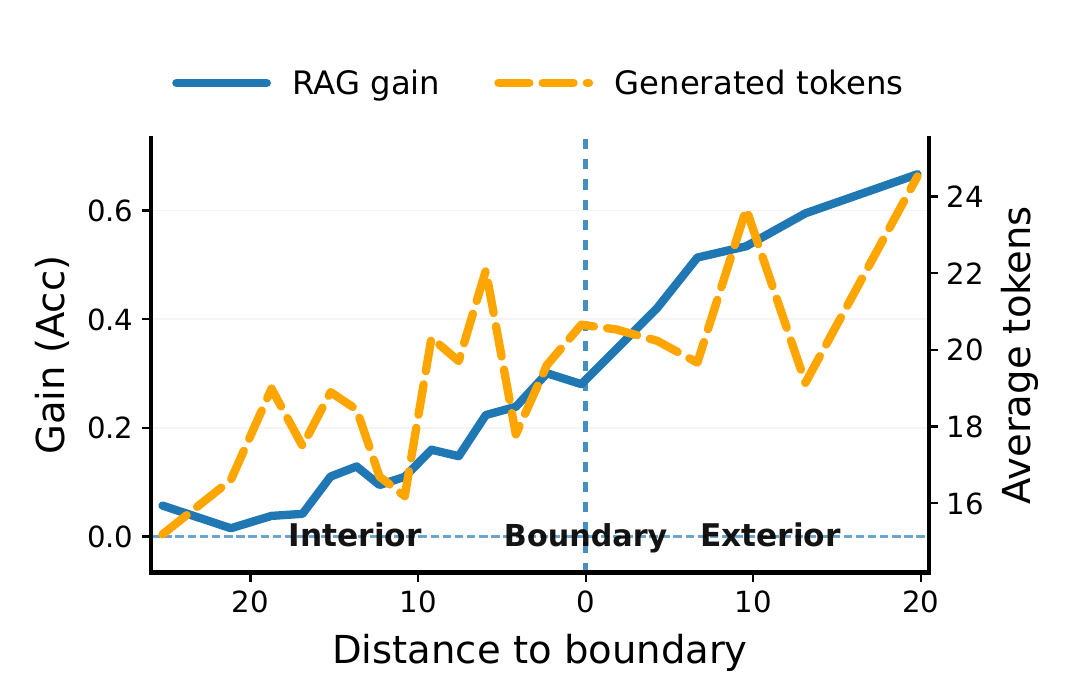}
        \caption{Qwen-14B: RAG}
        \label{fig:qwen_rag_gain}
    \end{subfigure}

    \caption{
    Intervention effects across the knowledge boundary on TriviaQA.
    }
    \label{fig:intervention_distance}
\end{figure*}

We study these interventions on TriviaQA, where factual recall provides a
relatively clean setting for examining their interaction with knowledge
accessibility. Figure~\ref{fig:intervention_distance} shows that these interventions are effective in different accessibility regions. Query rewriting yields its largest average gain in the interior, additional computation around the boundary, and external retrieval in the exterior. 
Experimental details are
provided in Appendix~\ref{app:intervention-experiments}.

\paragraph{Query rewriting is most useful when knowledge is already accessible.}
Query rewriting does not add new information or additional reasoning
computation. Instead, it changes how the query accesses knowledge already
stored in the model's parameters. On TriviaQA,
Figure~\ref{fig:llama_para_gain} shows that rewriting has its largest average
gain in the interior, where the required knowledge is more likely to be
parametrically accessible.

This suggests that some failures in the interior are from unsuccessful
access rather than missing knowledge. 
Paraphrasing the same query can provide
an alternative access path and recover the correct answer. As queries approach
the boundary, access becomes more sensitive to wording. Correctness varies more
across semantically equivalent paraphrases, and paraphrase stability decreases.
Farther into the exterior, rewriting becomes less effective because changing
the access path alone cannot compensate for increasingly unreliable parametric
access.

Prior work has shown that failure under one prompt does not necessarily imply
missing knowledge, since paraphrased prompts can recover correct responses
~\citep{jiang-etal-2020-know}. Our results further relate this phenomenon to
knowledge accessibility. Query rewriting is most effective when the required
knowledge is already within the model's accessible region, but its benefit
diminishes as parametric access weakens.

\paragraph{Additional computation is most useful near the boundary.}
Unlike query rewriting, CoT does not change the query provided to the
model. Instead, it allocates additional computation to accessing and using
knowledge already available from the model's parameters. 
On TriviaQA,
Figure~\ref{fig:llama_cot_gain} shows that CoT yields its largest average accuracy
gain near the knowledge boundary.

This pattern suggests that extra computation is most useful near the knowledge boundary, where the required knowledge is still accessible but access is unstable. Deep in the interior, direct access is already reliable, so there is little room for improvement. Near the boundary, the required knowledge is still available, but direct access is unstable. CoT gives the model more steps to retrieve and use this knowledge, so it has more room to improve than in the interior.
In the exterior, parametric access becomes unreliable. Since CoT adds computation but no new information, it cannot recover knowledge that the model cannot reliably access, so the gain decreases even as more tokens are generated.

Existing work improves reasoning efficiency and reduces inference cost by adapting compute to the amount of reasoning or thinking required by a query~\citep{wang-etal-2025-make,yong2025think,li-etal-2026-selfbudgeter}. Our findings complement this line of work by showing that knowledge accessibility can serve as a pre-generation signal for when additional computation is likely to help.

\paragraph{External retrieval is most useful outside the boundary.}
RAG differs from the previous interventions because it does not rely only on knowledge already available from the model's parameters. Instead, it supplies external information when parametric access is weak. On TriviaQA, Figure~\ref{fig:llama_rag_gain} shows that the average benefit of RAG increases with knowledge distance and is largest in the exterior. As queries move farther outside the accessible region, the required knowledge becomes less reliably available from the model's parameters, which makes external information increasingly useful.

This finding is closely related to adaptive RAG, which aims to decide when retrieval should be invoked at inference time. Existing methods make this decision using signals such as generation confidence, self-reflection, or query complexity~\citep{gao2023retrieval,jiang-etal-2023-active,jeong-etal-2024-adaptive}. Our results complement this line of work by showing that, in factual QA, the geometry of knowledge accessibility provides a pre-generation signal for when external retrieval is likely to help.

\paragraph{A unified pattern of accessibility interventions.}
The three interventions affect knowledge access in different ways, and their
benefits vary across the accessibility space.
In the \emph{interior}, where parametric access is reliable, query rewriting
can recover failures caused by how the query is expressed. Near the
\emph{boundary}, where access becomes unstable, additional reasoning yields its
largest gains. In the \emph{exterior}, where parametric access is
unreliable, external retrieval becomes more useful.

These results reveal a clear shift. As queries move from the \emph{interior} (reliable access), through the \emph{boundary} (unstable access), to the \emph{exterior} (unreliable access), the useful strategy shifts from query rewriting $\rightarrow$ additional computation $\rightarrow$ external retrieval.
Because knowledge distance is available before generation, this geometry may
provide a simple signal for deciding how a model should intervene on its own
inference process, which may help reduce inference cost~\citep{wang-etal-2026-x}.

\section{Related Work}
Prior work has studied the knowledge boundary of large language models, i.e.,
whether a model can distinguish queries it can answer from those it cannot~\citep{kadavath2022language, yin-etal-2023-large, yin-etal-2024-benchmarking, li-etal-2025-knowledge-boundary}. 
Several studies have further used LLM internal states to probe knowledge boundaries~\citep{slobodkin-etal-2023-curious,xiao-etal-2025-analyzing}.
Some of them estimate the boundary from the query alone before generation, which provides a useful signal for deciding how to answer a query~\citep{ni-etal-2025-towards, chen2026querylevel}.

However, these studies predict observed correctness directly, without decoupling whether errors are from limited knowledge recall or from failures in reasoning steps~\citep{press-etal-2023-measuring, li-goyal-2025-memorization, yang2026decoupling,cheang-etal-2026-llms, wang-etal-2026-x}.
More importantly, they focus mainly on whether knowledge boundary information is
encoded in hidden states, rather than how such information is organized geometrically in representation space.
In this work, we separate knowledge accessibility from reasoning and study
the geometry of knowledge accessibility in query representations.

\section{Conclusion}

This work shifts the study of the knowledge accessibility from whether
accessibility is encoded in LLM representations to how it is geometrically
organized. 
We find that accessible queries lie close to a common center. Accessibility drops as queries move farther from it.
It also tracks what
the model can access rather than whether it reasons correctly. 
Across datasets, the centers differ but
the accessibility ordering transfers. The geometry also indicates what kind
of help a query needs. Rewriting helps most in the interior, reasoning helps
most near the boundary, and retrieval helps most beyond it. Because this
signal is available before generation, it suggests that models could choose whether to answer directly, reason more, or retrieve information~\citep{chen2026querylevel}.

Our study has limitations that suggest several directions for future work. We
focus on single-fact queries and use correctness as a behavioral proxy for
accessibility. We do not separate knowledge that is never stored from
knowledge that is stored but not reached~\citep{allen-zhu2024physics}. Our evidence is also correlational.
Whether moving a representation toward the accessibility center can causally
impact recall remains an open question. Other open questions are how the
center emerges from pre-training~\citep{papyan2020prevalence} and whether the geometry can be extended
to reasoning capability.

\bibliography{iclr2027_conference}
\bibliographystyle{iclr2027_conference}

\appendix

\counterwithin{figure}{section}
\counterwithin{table}{section}
\numberwithin{equation}{section}

\section{Implementation Details}
\label{app:training_details}
\label{app:experimental_details}

\subsection{Representations and Correctness Labels}
\label{app:representations}

\paragraph{Query representations.}
For each query, the model receives the query alone, and we extract the
hidden state $\vh(x)$ at the final non-padding query token before any answer token is generated.

\paragraph{Standardization.}
Before fitting any geometry, each representation dimension is standardized
with the mean $\bm{\mu}$ and standard deviation $\bm{\sigma}$ computed on the
training split.
All geometries operate on standardized representations, so throughout the
paper $\vh(x)$ denotes the standardized representation.

\paragraph{Correctness labels.}
Correctness $c_{\pi}(x)$ (Eq.~\ref{eq:parametric-recall}) is decided by
\texttt{Mistral-Large}, used as a semantic-equivalence judge with temperature
zero. Given the query $x$, a response, and the reference answer set $\sY(x)$
including its aliases, the judge decides whether the response expresses one
of the reference answers. Unless stated otherwise, $\pi$ is greedy decoding.

\subsection{Geometry Models and Training}
\label{app:geometry_training}

The four accessibility scores are listed in
Table~\ref{tab:accessibility_hypotheses}. They are trained as follows.

\paragraph{Linear.}
Logistic regression with inverse regularization strength $C=1$ and the
L-BFGS solver.

\paragraph{MLP.}
One hidden layer with $m=128$ units, ReLU activation, and
dropout~\citep{srivastava2014dropout} of $0.1$. We use a batch size of 128 and
early stopping with a patience of 10 epochs based on validation AUROC.

\paragraph{Centered.}
The center $\vc$ is initialized to the mean representation of the correctly
answered training queries and trained for 600 epochs with a soft-margin
objective. The margin is selected from $\{5, 7, 10, 15, 20\}$ by validation
AUROC, separately for each layer and random seed.

\paragraph{A-Centered.}
A-Centered uses the distance in Eq.~\ref{eq:ellipsoidal-distance} with
$\mA$.  The
dimension weights are positive and have geometric mean one, which removes their
arbitrary global scale. Training has two stages. We first fit Centered and use
its learned center to initialize A-Centered. We then train A-Centered for 300
epochs. The margin is inherited from the
validation-selected Centered model.

\paragraph{Optimization.}
MLP, Centered, and A-Centered are trained with Adam~\citep{kingma2014adam}
using a learning rate of $10^{-2}$.

\paragraph{Data splits and seeds.}
The 40\%/10\%/50\% train/validation/test splits
(Section~\ref{sec:experiments}) are stratified by the correctness label. For
each seed, the same split is shared by all layers and all methods. The five
seeds are $\{42,43,44,45,46\}$.

\section{Details of the Geometry Analyses}
\label{app:geometry_analyses}

This section follows the order of Section~\ref{sec:experiments}.

\subsection{Testing the Centered Inductive Bias}
\label{app:centered_bias}

\begin{figure}[t]
    \centering
    \includegraphics[width=0.75\linewidth]{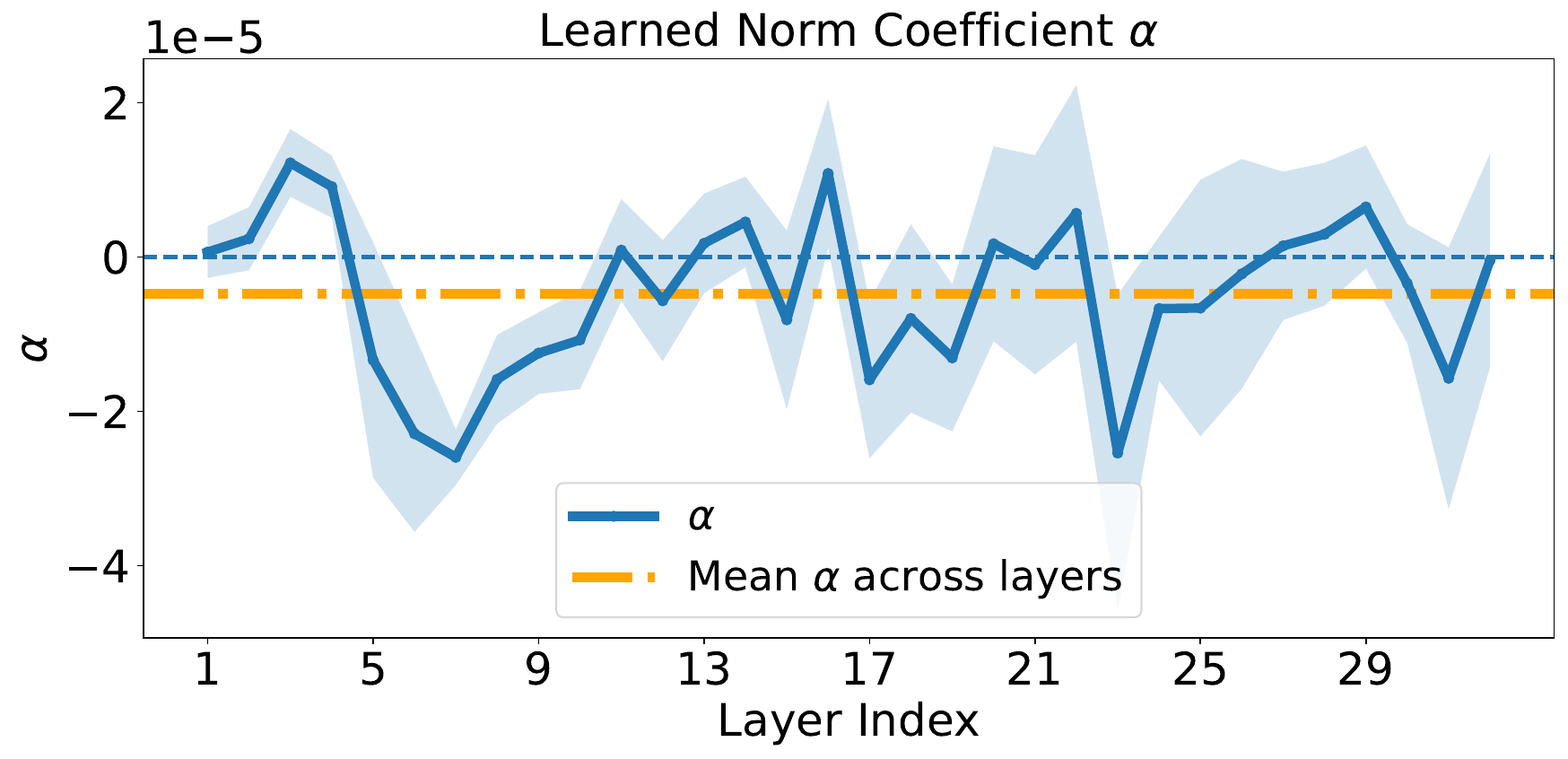}
    \caption{
    Learned norm coefficient $\alpha$ in Linear+Norm across layers on
    TriviaQA with Llama-8B. The line shows the mean over five random seeds
    and the shaded region shows $\pm 1$ standard deviation.
    }
    \label{fig:alpha_linear_norm}
\end{figure}

This section complements the Linear+Norm comparison in
Section~\ref{sec:geometry-comparison}. Linear+Norm uses the same splits as the
other geometries, and all preprocessing statistics are fitted on the training
split only. It is trained as a regularized logistic model, with the inverse
regularization strength selected from $C\in\{0.01,0.1,1,10\}$ by validation
AUROC.

Figure~\ref{fig:alpha_linear_norm} shows the learned coefficient $\alpha$
across layers. Although $\alpha$ varies across layers, its average is
negative. For $\alpha<0$, completing the square gives
\begin{equation}
\vw^\top\vh+\alpha\|\vh\|_2^2+b
=
-|\alpha|\,\bigl\|\vh-\tilde{\vc}\bigr\|_2^2+\mathrm{const},
\qquad
\tilde{\vc}=-\frac{\vw}{2\alpha},
\end{equation}
which is equivalent to the Centered score after positive rescaling, with an implied center
$\tilde{\vc}$. Thus, even without an explicit center, Linear+Norm tends to learn
a centered structure. Together with the stronger performance of Centered, this
supports the centered structure as a useful inductive bias rather than an
effect of greater model capacity.

\subsection{Controlled MQuAKE Evaluation}
\label{app:mquake_control}

We use MQuAKE to separate failures in knowledge accessibility from
failures in reasoning (Table~\ref{tab:mquake_boundary_position}). Each example
contains a multi-hop query together with the single-hop questions that probe
the facts required to answer it. We evaluate 1,000 examples with
\texttt{Llama-3.1-8B-Instruct}.

\paragraph{Behavioral groups.}
All single-hop questions and the multi-hop query are answered with the same
prompt and greedy decoding. Correct recall of all single-hop questions is used
as a behavioral proxy for accessible knowledge ($K$), and correctness of the
multi-hop query as reasoning success ($R$). Combining the two outcomes gives
four groups:
\emph{Success} (all single-hop questions and the multi-hop query correct),
\emph{Reasoning Failure} (all single-hop questions correct, multi-hop query
incorrect),
\emph{Knowledge Failure} (at least one single-hop question and the multi-hop
query incorrect), and
\emph{Recovered Success} (at least one single-hop question incorrect, multi-hop
query correct).

\paragraph{Frozen TriviaQA geometry.}
No geometry is fitted on MQuAKE. We freeze the A-Centered geometry learned on
TriviaQA. Each canonical MQuAKE query is
represented exactly as in Appendix~\ref{app:representations} at the selected
TriviaQA layer, and its distance is
computed under the frozen geometry.

\subsection{Behavior of Knowledge Accessibility}
\label{app:boundary_instability}

\begin{figure*}[t]
    \centering
    \begin{subfigure}[t]{0.25\textwidth}
        \centering
        \includegraphics[width=\linewidth]
        {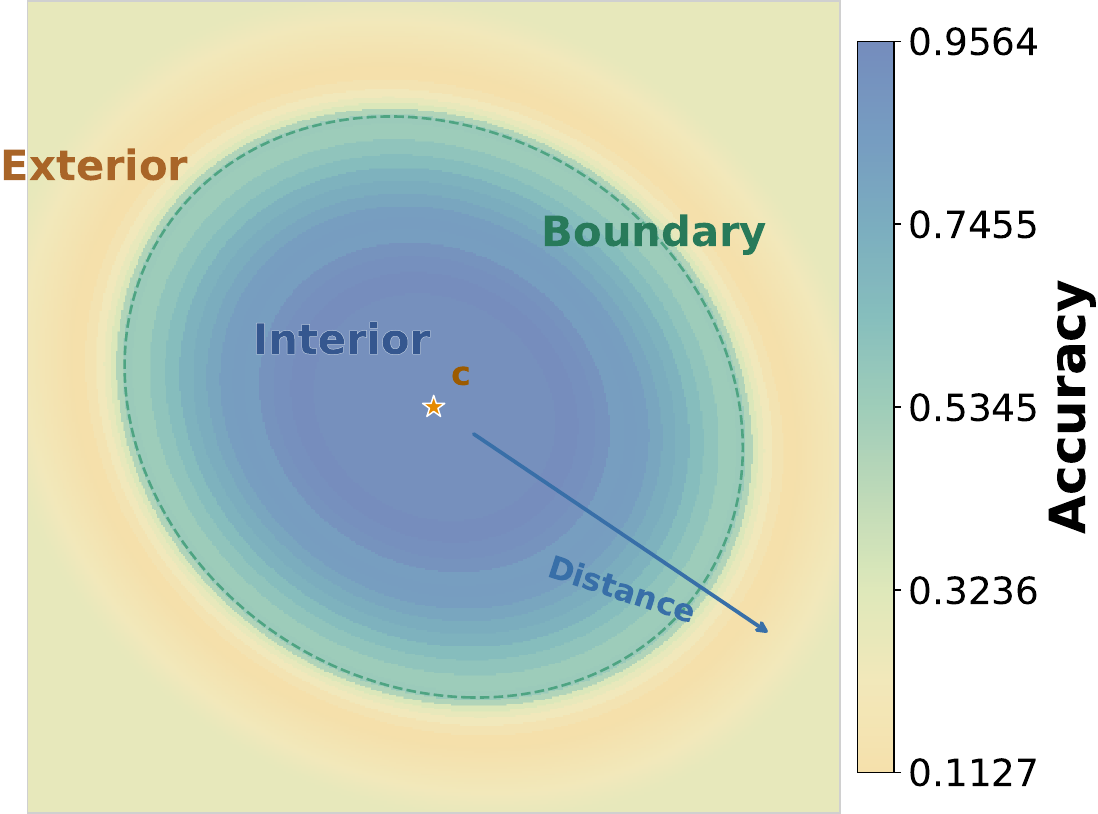}
        \caption{Accuracy vs. distance}
        \label{fig:qwen14b_knowledge_landscape}
    \end{subfigure}
    \hfill
    \begin{subfigure}[t]{0.3\textwidth}
        \centering
        \includegraphics[width=\linewidth]
        {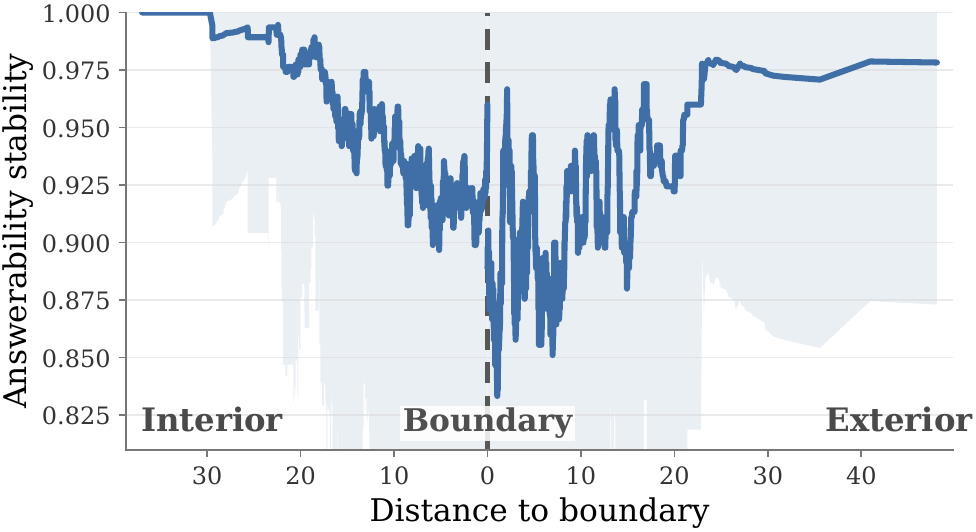}
        \caption{Correctness stability}
        \label{fig:qwen14b_answerability_stability}
    \end{subfigure}
    \hfill
    \begin{subfigure}[t]{0.3\textwidth}
        \centering
        \includegraphics[width=\linewidth]
        {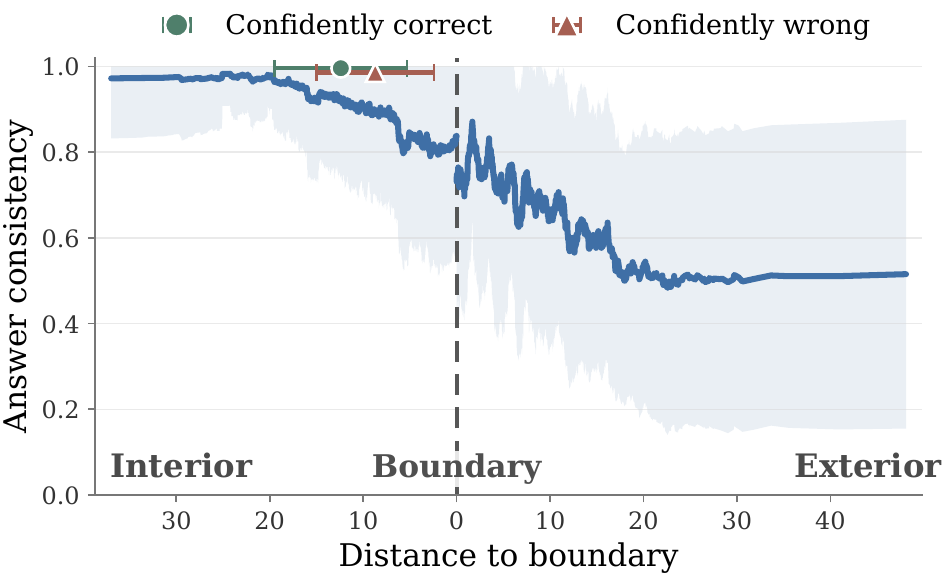}
        \caption{Answer consistency}
        \label{fig:qwen14b_answer_consistency}
    \end{subfigure}
    \caption{
    Behavior of knowledge accessibility for Qwen-14B on TriviaQA
    (counterpart of Figure~\ref{fig:knowledge_boundary_properties}).
    (a) Accessibility decreases with distance from the center.
    (b) Correctness is least stable near the boundary.
    (c) Consistency decreases with distance from the center.
    }
    \label{fig:qwen14b_knowledge_boundary_properties}
\end{figure*}

This section gives the details for Section~\ref{sec:boundary_behavior}; the
results for Qwen-14B are shown in
Figure~\ref{fig:qwen14b_knowledge_boundary_properties}.

\paragraph{Frozen geometry on held-out queries.}
All analyses use test queries only. Their distances are computed under the geometry learned on the training
split.

\paragraph{Repeated sampling and correctness stability.}
For each test query, we draw $N$ independent responses under a fixed
stochastic decoding strategy $\pi$. Each response is judged independently as in
Appendix~\ref{app:representations}, which gives $c_{\pi}^{(i)}(x)$. The
empirical correctness probability $\hat{p}_{\pi}(x)$ and the correctness
stability $S_{\pi}(x)$ then follow Eq.~\ref{eq:correctness-stability}. 

\paragraph{Answer consistency.}
Answer consistency measures whether repeated generations produce the same
surface-form answer, regardless of whether that answer is correct.  For
visualization, queries with $\kappa_{\pi}(x)\geq 0.9$ are marked as
\emph{confidently correct} if their modal normalized answer is judged correct,
and as \emph{confidently wrong} otherwise. These categories are only used for
visualization.

\subsection{Cross-Dataset Transfer and Dataset-Specific Geometry}
\label{app:boundary-transfer}

\begin{figure}[t]
\centering
\includegraphics[width=0.88\linewidth]{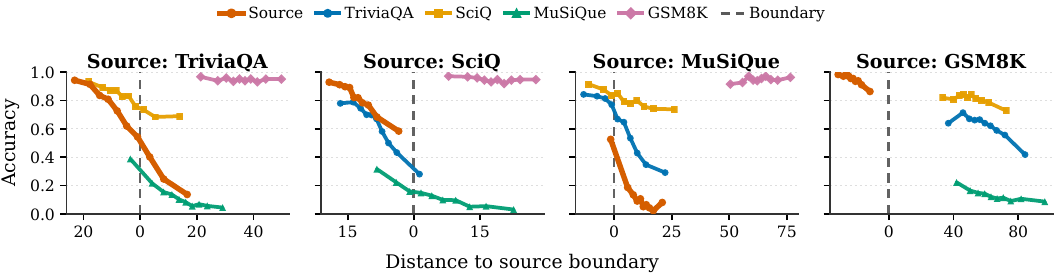}
\caption{
Transferability of the distance and accuracy relation for Qwen-14B
(counterpart of Figure~\ref{fig:cross-dataset-transfer}). Accuracy generally
decreases with distance across target datasets.
}
\label{fig:qwen14b_cross-dataset-transfer}
\end{figure}

\begin{figure}[t]
\centering
\includegraphics[width=0.9\linewidth]{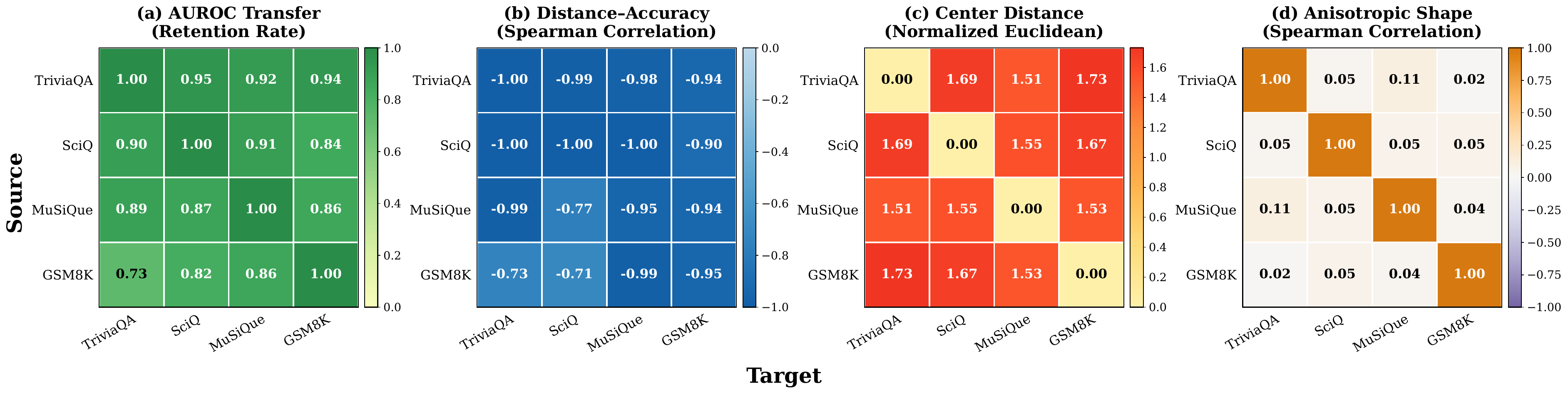}
\caption{
Cross-dataset diagnostics for Llama-8B (layer 18).
(a) AUROC retention, (b) distance--accuracy correlation,
(c) distance between learned centers, and
(d) correlation between learned dimension weights.
The distance-based ordering transfers across datasets even though the centers
and dimension weights are dataset-specific.
}
\label{fig:geometry_heatmaps}
\end{figure}

This section gives the details for Section~\ref{sec:geometric-transferability}.
We separate two questions: whether the \emph{relation} between distance and
correctness transfers across datasets, and whether the \emph{geometric
parameters} themselves (center and dimension weights) are shared.

\paragraph{Transfer protocol.}
We use TriviaQA, SciQ, MuSiQue, and GSM8K, with the best-performing layer selected for Llama-8B. Each dataset is used in turn as the source $s$. Its A-Centered
geometry is
applied without relearning. 

\paragraph{Distance--accuracy relation.}
Queries of a target dataset $t$ are grouped into distance bins $B_k$ under the
source geometry. For each bin we compute the mean distance and the accuracy,
\begin{equation}
\bar{D}_k=\frac{1}{|B_k|}\sum_{x\in B_k}D_s(x),
\qquad
\mathrm{Acc}(B_k)=\frac{1}{|B_k|}\sum_{x\in B_k}c(x),
\end{equation}
which gives the curves in Figures~\ref{fig:cross-dataset-transfer}
and~\ref{fig:qwen14b_cross-dataset-transfer}. The relation is summarized by
the Spearman correlation
\begin{equation}
\rho_{s\rightarrow t}
=
\rho_{\mathrm{Spearman}}\bigl(\{\bar{D}_k\}_k,\{\mathrm{Acc}(B_k)\}_k\bigr).
\label{eq:transfer-spearman}
\end{equation}
A negative value means that target queries farther from the source center are
answered less accurately. This only requires the source geometry to preserve
the relative ordering of target queries, not to share their center or shape.

\paragraph{Cross-dataset diagnostics.}
Figure~\ref{fig:geometry_heatmaps} reports four pairwise matrices. In (a) and
(b), rows are source datasets and columns are target datasets. In (c) and (d),
entries compare the geometries learned independently on each pair of datasets.
\begin{itemize}
\item[(a)] \textbf{AUROC retention.} The transfer AUROC is normalized by the
AUROC of the target's own geometry,
$\mathrm{Retention}_{s\rightarrow t}
=\mathrm{AUROC}_{s\rightarrow t}/\mathrm{AUROC}_{t\rightarrow t}$.
\item[(b)] \textbf{Distance--accuracy monotonicity.} Entry $(s,t)$ is
$\rho_{s\rightarrow t}$ from Eq.~\ref{eq:transfer-spearman}.
\item[(c)] \textbf{Center variation.} Since each geometry is fitted after
dataset-specific standardization, we first map the learned centers back to the
original representation space. We then measure the distance between centers and
normalize it by the overall scale of the representations. Smaller values
indicate more similar centers.

\item[(d)] \textbf{Dimension-weight variation.} We similarly map the learned
dimension weights back to the original representation space and compare them
using Spearman correlation. A high correlation means that the two geometries
emphasize similar dimensions.
\end{itemize}

\paragraph{Interpretation.}
Panels (a) and (b) show that transferred geometries retain much of the
target-specific AUROC and that distance--accuracy correlations remain negative
for all source--target pairs. Panels (c) and (d) show that the learned centers
and dimension weights differ across datasets. 

\section{Intervention Experiments}
\label{app:intervention-experiments}

This section provides the experimental details for
Section~\ref{sec:model-behavior}.

\subsection{Shared Setup}
\label{app:intervention-setup}

\paragraph{Geometry and queries.}
All interventions are evaluated on the TriviaQA test set. We use the same
frozen A-Centered geometry and the same selected layer as in the main
experiments.

\paragraph{Evaluation.}
For each query, we compare the correctness of the original greedy answer with
the correctness after an intervention. Correctness is judged as in
Appendix~\ref{app:representations}. For CoT and RAG, only the final extracted
answer is given to the judge, without the reasoning trace or retrieved context.

\paragraph{Binning.}
Queries are grouped into 20 equal-frequency bins according to their signed
distance to the learned boundary. Negative distance corresponds to the
interior side, while positive distance corresponds to the exterior side.
All curves report raw bin-level statistics without smoothing or interpolation.

\subsection{Query Rewriting}
\label{app:paraphrase-intervention}

\paragraph{Paraphrase generation.}
We use \texttt{Mistral-Large} with temperature $0.8$ to generate five
semantically equivalent paraphrases for each query. The paraphrases are
instructed to preserve the original meaning, while changing wording and sentence structure.

\paragraph{Answering and evaluation.}
The model answers each paraphrase independently using only its internal
knowledge. We use greedy decoding with at most 128 new tokens and do not use
chain-of-thought or retrieval. We keep queries for which all five paraphrases
receive valid correctness judgments.

For each original query, we compute the fraction of paraphrases answered
correctly. We also measure paraphrase stability from the consistency of the
five correctness outcomes. The rewriting gain reported in
Figure~\ref{fig:intervention_distance} is the change in correctness relative
to the original query.

\subsection{Chain-of-Thought}
\label{app:cot-intervention}

The model is prompted to reason step by step using only its internal knowledge.
We allow at most five short reasoning steps before the final answer. Generation
uses greedy decoding with at most 2,048 new tokens.

This intervention adds computation but no external information. We record both
answer correctness and the number of generated tokens, which allows us to
compare the benefit and cost of additional reasoning across different
knowledge-boundary regions.

\subsection{Retrieval-Augmented Generation}
\label{app:rag-intervention}

For retrieval, we use the evidence provided by TriviaQA. Each test question is
matched to its corresponding example in the TriviaQA \texttt{rc} validation
split. We use the highest-ranked non-empty \texttt{search\_context} and truncate
it to at most 4,096 tokens.

The retrieved context is prepended to the question, and the model is instructed
to answer using it as external evidence. We do not request chain-of-thought so
that the intervention mainly reflects the effect of additional information
rather than additional reasoning. Generation uses greedy decoding with at most
256 new tokens.

Queries without a matching TriviaQA example or usable retrieval context are
excluded from the RAG analysis.


\end{document}